\documentclass{article}

    \PassOptionsToPackage{numbers, compress}{natbib}
 \usepackage[preprint]{neurips_2026}

\usepackage[utf8]{inputenc} 
\usepackage[T1]{fontenc}    
\usepackage{hyperref}       
\usepackage{url}            
\usepackage{booktabs}       
\usepackage{amsfonts}       
\usepackage{nicefrac}       
\usepackage{microtype}      
\usepackage{xcolor}         

\usepackage{pifont} 
\usepackage[dvipsnames]{xcolor}
\newcommand{\cmark}{\ding{51}}%
\newcommand{\xmark}{\ding{55}}%

\usepackage{booktabs}   
\usepackage{multirow}   
\usepackage{graphicx}   
\usepackage{amssymb}    

\definecolor{gold}{RGB}{255, 245, 200}
\definecolor{lightgray}{gray}{0.92}
\usepackage[table]{xcolor}
\definecolor{green}{HTML}{32CD32} 

\definecolor{q_color}{RGB}{25, 72, 128}  
\definecolor{a_color}{RGB}{150, 0, 0}    

\usepackage{amsmath}    
\usepackage{array}      
\usepackage[table]{xcolor}
\usepackage{subcaption}
\usepackage{wrapfig}

\title{Decoupled Optimization for Teacher--Student Semi-Supervised Learning via a Pioneer Student}

\author{%
  Haorong Han \\
  School of Computer Science and Technology \\
  Beijing Jiaotong University \\
  \texttt{harohan@bjtu.edu.cn} \\
  \And
  Jidong Yuan\thanks{Corresponding author.} \\
  School of Computer Science and Technology \\
  Beijing Jiaotong University \\
  \texttt{yuanjd@bjtu.edu.cn} \\
  \AND
  Chixuan Wei \\
  School of Data Science and Intelligent Media \\
  Communication University of China \\
  \texttt{chxwei@cuc.edu.cn} \\
  \And
  Yongqi Sun \\
  School of Computer Science and Technology \\
  Beijing Jiaotong University \\
  \texttt{yqsun@bjtu.edu.cn} \\
}

\begin{document}

\maketitle

\begin{abstract}
Semi-supervised learning (SSL) relies on two core mechanisms: self-training under the Teacher--Student (T--S) framework and joint optimization of labeled and unlabeled losses. Despite their effectiveness, we find both mechanisms introduce distinct optimization pathologies. First, parameter coupling enforces strict synchronization between teacher and student, where strong regularization on the student degrades the teacher's fitting ability, thereby limiting the permissible generalization intensity. Second, the imbalance in gradient update consistency between labeled and unlabeled losses drives the shared parameters to prematurely converge to labeled-dominated local minima, creating a bottleneck for global optimization. To address both issues, we propose the Pioneer Student (PiS), an auxiliary branch that operates in an independent parameter space and periodically transfers accumulated knowledge back to the T--S model. Extensive experiments show that PiS is a universal plug-and-play module that consistently improves mainstream SSL methods. Code is available at \url{https://github.com/hhrd9/PiS}.
\end{abstract}

\section{Introduction}

Semi-supervised learning (SSL) aims to leverage unlabeled data to improve model performance under limited supervision~\cite{lee2013pseudo,rasmus2015semi,miyato2018virtual,sohn2020fixmatch}. It inherently requires strong generalization, not only to extrapolate to unseen data as in supervised learning~\cite{he2016deep}, but more crucially because its training relies heavily on noisy pseudo-labels.
Intuitively, increasing the strength of data augmentation should improve robustness and generalization.
However, empirical observations in SSL reveal a puzzling phenomenon: as shown in Table~\ref{tab:intro_motivation}, while moderate augmentation improves performance, further increasing its strength leads to a clear degradation.
For instance, under RandAugment~\cite{cubuk2020randaugment}, increasing its hyperparameters (i.e., from (2,3) to (3,5), where larger values correspond to stronger transformations) improves performance, whereas pushing them to (5,10) results in a noticeable drop.

Is this the fundamental limit of generalization capacity in SSL?
Interestingly, when pseudo-labels pre-recorded offline from a complete run under the (3,5) setting are used as exogenous supervision to train a model from scratch under the (5,10) setting (with identical data ordering ensured by fixing the random seed), performance can be further improved.
This reveals that the bottleneck lies not in the model's capacity to withstand strong augmentations, but in the pseudo-labeling mechanism.

To understand this discrepancy, we undertake a systematic investigation into the optimization mechanics of the widely adopted Teacher--Student (T--S) framework.
Through our analysis (Section~\ref{sec:fgd}), we identify that the core issue stems from the parameter coupling inherent in the self-training mechanism~\cite{sohn2020fixmatch}.
We term this the \textbf{Fitting--Generalization Dilemma (FGD)}: the student model's pursuit of strong generalization inadvertently propagates through the coupled parameters to degrade the teacher model's fitting fidelity on the training data.
This degradation lowers the quality of the generated pseudo-labels, thereby establishing a hard ceiling on the permissible generalization intensity.

\begin{wraptable}{r}{0.54\textwidth}
    \centering
    \vspace{-0pt} 
    \caption{\textbf{Performance of FreeMatch~\cite{wang2023freematch} on CIFAR-100~\cite{krizhevsky2009learning} (2 labels/class) under varying augmentation intensities.} Online pseudo-labels generated by the synchronized teacher are endogenous, whereas fixed ones pre-recorded offline (e.g., from an RA(3,5) run) are exogenous.}

    \label{tab:intro_motivation}
    \resizebox{0.54\textwidth}{!}{
        \begin{tabular}{llcc}
            \toprule
            Supervision Type & Augmentation & Acc. (\%) & \textcolor{gray!60}{Pseudo-Label Acc. (\%)} \\
            \midrule
            \multirow{3}{*}{Endogenous} & RA(2,3) & 78.13 & \textcolor{gray!60}{77.97} \\
                                        & RA(3,5) & 78.46 & \textcolor{gray!60}{77.92} \\
                                        & RA(5,10) & 76.27 & \textcolor{gray!60}{75.25} \\
            \midrule
            Exogenous & RA(5,10) & 78.80 & \textcolor{gray!60}{77.92} \\
            \bottomrule
        \end{tabular}
    }
    \vspace{-10pt}
\end{wraptable}

Beyond the FGD, it is a standard consensus in SSL to jointly optimize the labeled and unlabeled losses within the same parameter space~\cite{sohn2020fixmatch,wang2023freematch,nguyen2024sequencematch,han2025regmixmatch}.
However, our investigation (Section~\ref{sec:ldt}) reveals that this formulation conceals a deeper optimization conflict, which we conceptualize as the \textbf{Label-Driven Trap (LDT)}.
Put simply, by introducing an \emph{Update Consistency} metric, we observe that the gradient updates driven by the scarce labeled data are highly directed, whereas the unlabeled updates are exploratory.
This asymmetry drives the shared parameters to prematurely converge to local minima that favor the labeled loss but are suboptimal for optimizing unlabeled data, thereby creating a bottleneck for global optimization.

Based on our interpretations, it is intriguing that solving these two seemingly unrelated pathologies points toward the same architectural necessity: an independent parameter space.
Consequently, we propose the \textbf{Pioneer Student (PiS)}, an auxiliary branch whose parameters are completely decoupled from the T--S model.
Operating in this isolated parameter space, PiS achieves two forms of decoupled optimization simultaneously: it allows the student to explore extreme generalization without compromising the teacher's fitting fidelity, and it enables the autonomous optimization of the unlabeled loss, free from the interference of the labeled data.
Equally important, we introduce the periodic parameter overwrite that teleports the accumulated robust knowledge from PiS back into the T--S model, ensuring that the overall framework achieves stage-wise performance breakthroughs and maximizes the utility of the pioneer branch. We summarize our contributions as follows:

\begin{itemize}
    \item We analyze and identify two fundamental optimization coupling pathologies in conventional SSL practices: the FGD and the LDT, which restrict the generalization upper bound and create a global optimization bottleneck, respectively.
    \item We propose PiS, a simple yet highly effective dual-branch strategy that simultaneously resolves both the FGD and the LDT through optimization decoupling.
    \item As a plug-and-play module, we demonstrate through extensive experiments that PiS significantly enhances existing mainstream SSL methods and exhibits superior practical potential.
\end{itemize}

\section{Related Work}

\textbf{Teacher--Student framework} has long been intertwined with the evolution of SSL. The pioneering work Pseudo-labeling~\cite{lee2013pseudo} establishes the fundamental logic of self-training by converting model predictions on unlabeled data into one-hot labels to supervise a parameter-sharing student network. Subsequent research has evolved along two directions: on the teacher side, Temporal Ensembling~\cite{laine2017temporal} improves prediction stability by aggregating historical predictions, while Mean Teacher~\cite{tarvainen2017mean} introduces EMA of weights to construct a more robust teacher. On the student side, VAT~\cite{miyato2018virtual} creates challenging input conditions through adversarial perturbations, and Noisy Student~\cite{noisystud} significantly pushes the generalization upper bound by injecting diverse noise strategies.

\textbf{Consistency regularization with weak--strong augmentations} has revitalized the T--S framework, where the teacher generates pseudo-labels from weakly-augmented views while the student learns under heavy perturbations. Building upon this, FixMatch~\cite{sohn2020fixmatch} and its derivatives~\cite{zhang2021flexmatch,wang2023freematch,chen2023softmatch} introduce fixed or dynamic thresholds to filter low-confidence predictions to ensure teacher supervision fidelity. Other efforts~\cite{li2021comatch,nguyen2024sequencematch,zhang2025towards} improve student regularization via multiple augmentation paths or auxiliary constraints. Recently, RegMixMatch~\cite{han2025regmixmatch}, related to ours, identifies that the high-entropy behavior induced by Mixup~\cite{zhang2018mixup} can degrade teacher confidence and proposes refinements building upon prior approaches~\cite{berthelot2019mixmatch,berthelot2019remixmatch}. Transcending the refinement of specific generalization techniques, we elucidate and reconcile the deeper FGD inherent in the T--S framework, enabling universal performance gains across mainstream SSL methods.


\textbf{Labeled data's optimization role} is increasingly scrutinized. MPL~\cite{pham2021meta} leverages model's learning on labeled data to update teacher's parameters, guiding pseudo-label generation. Closely related to our study, FlatMatch~\cite{huang2023flatmatch} empirically identifies rapid labeled loss convergence hinders global optimization, and applies cross-sharpness penalties as mitigation. However, it neither analyzes underlying mechanism nor quantifies its impact on SSL performance, while labeled and unlabeled optimization remain entangled within shared parameters. In contrast, we formalize this pathology as the LDT, directly demonstrate its adverse effect on final accuracy (Section~\ref{sec:dualb}), and propose PiS as a structural resolution that fully decouples the two optimization into independent parameter spaces.


\textbf{Dual-branch methods} have also been explored in SSL and beyond. In SSL, co-training-based methods~\cite{co-training,ke2019dual,rothenberger2023meta,huang2026bidirectional} employ two parallel T--S models to stabilize training. However, each branch still performs joint loss optimization and maintains parameter coupling, leaving both FGD and LDT unresolved while incurring roughly doubled computational overhead. Beyond SSL, the Lookahead optimizer~\cite{zhang2019lookahead} updates slow weights by \textit{interpolating} toward fast weights to obtain a smoother parameter state, whereas our approach employs hard \textit{overwrites} to replace the T-S weights with those of the PiS, forcefully teleporting the model away from labeled-dominated local minima.

\section{Coupled Optimization Pathologies in Semi-Supervised Learning}
\label{sec:pathologies}

In this section, we systematically dissect the optimization dynamics of the widely adopted T--S framework to uncover two fundamental optimization coupling pathologies: the FGD and the LDT.

\subsection{\textcolor{q_color}{Q1:} Why Does SSL Degrade Under Strong Generalization?}
\label{sec:fgd}

\textbf{The T--S Framework and Self-Training.}
The cornerstone of modern SSL is the T--S paradigm powered by weak--strong augmentations~\cite{sohn2020fixmatch}. Given an unlabeled sample $x_u$, the student $f_{\theta_s}$ produces a prediction $p_s = f_{\theta_s}(\mathcal{A}(x_u))$ on its strongly augmented version $\mathcal{A}(x_u)$, while the teacher $f_{\theta_t}$ produces $p_t = f_{\theta_t}(\alpha(x_u))$ on the weakly augmented version $\alpha(x_u)$, from which the pseudo-label $q_u = \arg\max(p_t)$ is derived. The student is optimized via the consistency loss:
\begin{equation}
    \mathcal{L}_u(x_u, q_u; \theta_s) = \mathcal{H}(q_u, \sigma(p_s)),
\label{equ:lu}
\end{equation}
where $\sigma(\cdot)$ denotes the softmax operation and $\mathcal{H}(\cdot,\cdot)$ is the cross-entropy loss. Crucially, to form a self-training loop, teacher's parameters $\theta_t$ strictly track student's parameters $\theta_s$ via direct duplication ($\theta_t \leftarrow \theta_s$) or an EMA update ($\theta_t \leftarrow m \theta_t + (1 - m)\theta_s$). This parameter coupling allows the teacher to inherit the student's evolving representations, continuously refining the pseudo-label quality.

\textbf{The Cost of Parameter Coupling.}
This elegant self-training mechanism, however, conceals a structural flaw. Unlike supervised learning with fixed ground-truth labels, the supervision signal $q_u$ here is an endogenous variable determined entirely by $\theta_t$. Because $\theta_t$ is passively synchronized with $\theta_s$, the single optimization objective $\min_{\theta_s}\mathcal{L}_u$ implicitly binds two conflicting demands: the teacher must perfectly \emph{fit} the training data to yield reliable $q_u$, whereas the student must aggressively \emph{generalize} to withstand strong augmentations $\mathcal{A}$.

Since fitting and generalization represent opposing optimization directions, enforcing strict parameter synchronization creates a zero-sum game. As shown in Table~\ref{tab:intro_motivation}, strong generalization strategies significantly degrade pseudo-label accuracy, ultimately hurting performance. Conversely, when the teacher is relieved from pseudo-label generation and exogenous pseudo-labels obtained under moderate generalization are employed, a student trained with strong generalization strategies achieves better performance.

\textbf{\textcolor{a_color}{A1:} The Fitting--Generalization Dilemma (FGD).}
Within the T--S framework, the supervision $q_u$ is not an external ground truth but an endogenous variable that depends on $\theta_s$ through the synchronized teacher, leading to the objective $\min_{\theta_s} \mathcal{L}_u\!\left(x_u,\, q_u(\theta_s);\, \theta_s\right)$, where a single parameter state $\theta_s$ is forced to simultaneously act as the optimizer of the loss (demanding generalization) and the source of the supervision target $q_u(\theta_s)$ (demanding fitting). The student's pursuit of stronger generalization inevitably degrades the teacher's fitting fidelity, compromising pseudo-label reliability and establishing a hard ceiling on the permissible generalization intensity.

\subsection{\textcolor{q_color}{Q2:} What Are the Consequences of Jointly Optimizing Both Losses?}
\label{sec:ldt}

\textbf{Joint Loss Optimization.}
SSL methods conventionally optimize both losses within the same parameter space. Given a labeled sample $x_l \in \mathcal{X}_l$ with ground-truth label $y_l$, the labeled loss is:
\begin{equation}
    \mathcal{L}_l(x_l, y_l; \theta_s) = \mathcal{H}(y_l, \sigma(f_{\theta_s}(x_l))).
\label{equ:ll}
\end{equation}
The standard SSL objective then becomes:
\begin{equation}
    \min_{\theta_s}\; \mathcal{L}_l(x_l, y_l;\, \theta_s) + \mathcal{L}_u(x_u, q_u;\, \theta_s).
\label{equ:joint}
\end{equation}

\begin{wrapfigure}{r}{0.48\textwidth} 
  \vspace{-12pt} 
  \centering
  \captionsetup[subfigure]{font=scriptsize,labelfont=scriptsize}
  
  \begin{subfigure}{0.48\linewidth}
    \centering
    \includegraphics[width=\linewidth]{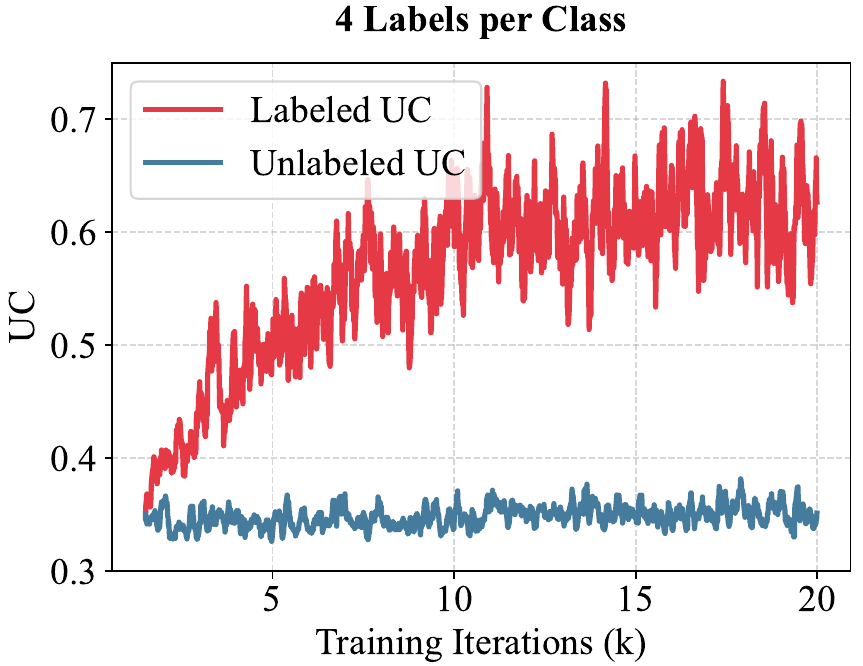}
  \end{subfigure}
  \hfill
  \begin{subfigure}{0.48\linewidth}
    \centering
    \includegraphics[width=\linewidth]{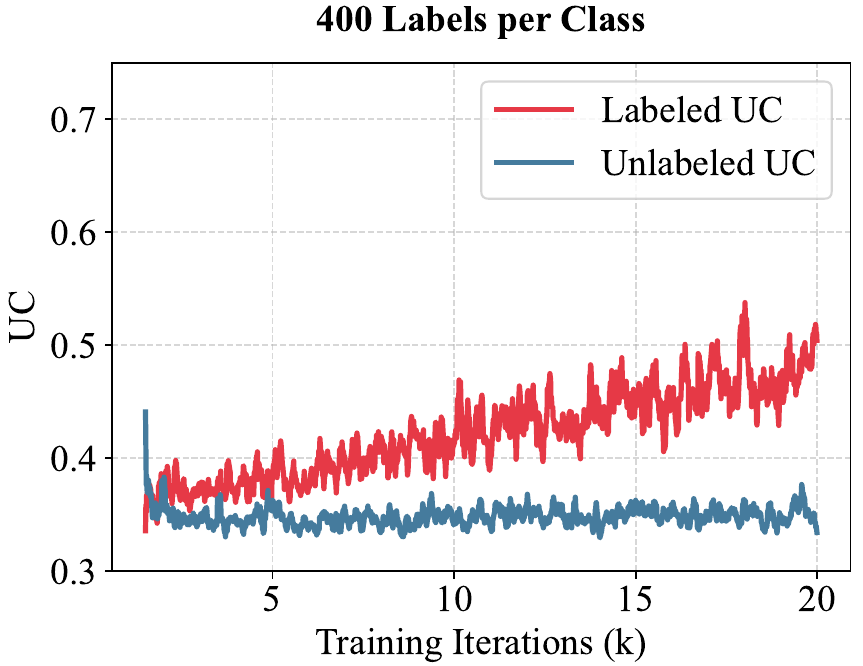}
  \end{subfigure}
  
  \caption{\textbf{UC of labeled and unlabeled losses.}}
  \label{fig:UC}
  \vspace{-14pt}
\end{wrapfigure}

\textbf{Update Consistency.}
To characterize the optimization dynamics, we introduce \emph{Update Consistency} (UC), which measures the directional coherence of $k$ consecutive gradient updates:
\begin{equation}
    \text{UC} = \frac{\left\lVert\sum_{i=1}^k g^{(i)}\right\rVert}{\sum_{i=1}^k \left\lVert g^{(i)}\right\rVert}.
\label{equ:uc}
\end{equation}
In our experiments, we set $k=10$. A high UC indicates that gradients consistently point in the same direction, reflecting a directed and dominant optimization signal; a low UC indicates exploratory, diffuse updates.
Figure~\ref{fig:UC} reports the UC of labeled and unlabeled losses in FreeMatch~\cite{wang2023freematch} on STL-10~\cite{coates2011analysis} under 4-label and 400-label settings per class \textit{during early training}. We observe that the labeled loss exhibits higher UC than the unlabeled loss, with the gap becoming more pronounced in the 4-label setting. This suggests that, under label scarcity, limited label supervision induces a more directed optimization path than the unlabeled loss, which is optimized over large-scale unlabeled data and thus remains exploratory.

\begin{figure*}[t]
    \centering
    \includegraphics[width=0.90\textwidth]{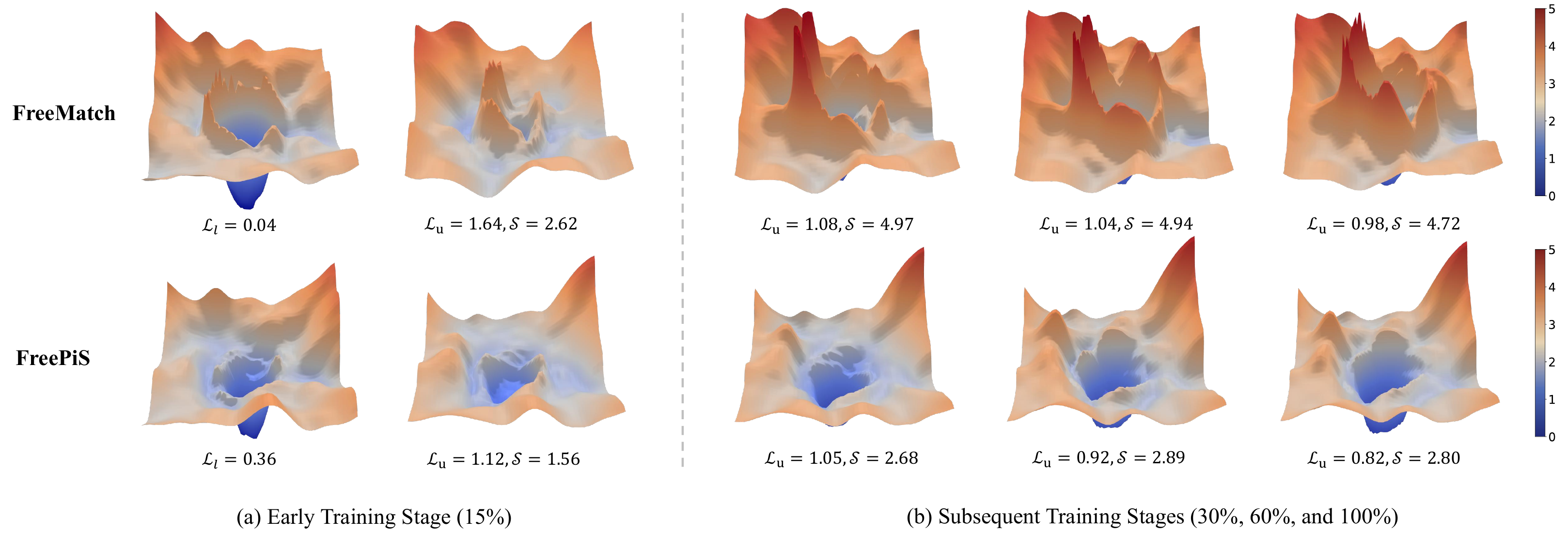}
    \caption{\textbf{Visualization of loss landscapes: FreeMatch vs ours (FreePiS).} 
    Landscapes are obtained by perturbing the model parameters (landscape origin) along two orthogonal directions. $\mathcal{L}_l$ and $\mathcal{L}_u$ denote the labeled and unlabeled loss values at the origin, respectively, while $\mathcal{S}$ measures landscape sharpness, which reflects generalization potential (see Appendix~\ref{appendix:sharpness} for formal definition).}
    \label{fig:losscape}
    \vspace{-5pt}
\end{figure*}

\textbf{Loss Landscape Analysis.}
To examine the effect of this early-stage UC imbalance within a shared parameter space, we visualize the loss landscape~\cite{li2018visualizing} under the 4-label setting (first row of Figure~\ref{fig:losscape}).
Three key observations emerge.
First, the labeled loss rapidly converges to a near-optimal state ($\mathcal{L}_l = 0.04$), consistent with its high UC. 
Second, the unlabeled loss landscape closely follows the geometry of the labeled counterpart due to shared parameters, yet remains severely under-optimized ($\mathcal{L}_u = 1.64$) and exhibits significantly higher sharpness, portending poor generalization potential~\cite{neyshabur2017exploring}.
Third, the unlabeled loss landscapes remains geometrically consistent while maintaining high sharpness throughout subsequent training stages ($\mathcal{S} \in [4.7, 5.0]$), indicating that parameter state shaped in early training permanently constrains the optimization trajectory.

\textbf{\textcolor{a_color}{A2:} The Label-Driven Trap (LDT).}
Under label scarcity, the two gradient signals in Eq.~\ref{equ:joint} are highly imbalanced, with
$\text{UC}\!\left(\nabla_{\theta_s}\mathcal{L}_l\right) \gg \text{UC}\!\left(\nabla_{\theta_s}\mathcal{L}_u\right)$, where the highly directed labeled gradient disproportionately biases the early optimization trajectory (see Appendix~\ref{appendix:grad_norm} for gradient magnitude analysis), steering the shared parameters toward a local basin defined by the scarce labeled data. As this basin is shaped by \textit{limited} labeled samples, it is non-global and suboptimal for minimizing the unlabeled loss. More importantly, once the parameters are trapped in this basin, the unlabeled loss can only be optimized within its neighborhood, creating a bottleneck for global optimization.

\section{Method}
\label{sec:Method}

\begin{figure*}[t]
    \centering
    \includegraphics[width=0.95\textwidth]{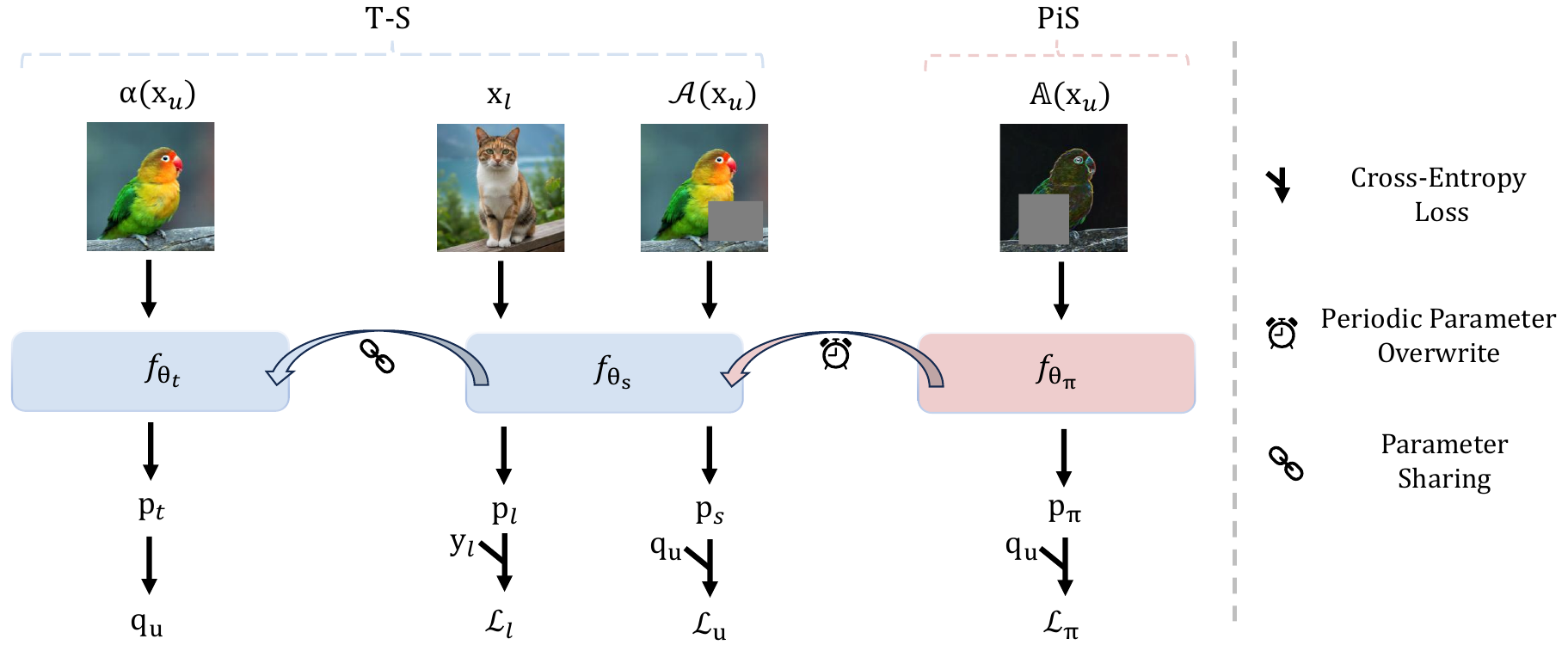}
    \caption{\textbf{Overview of T--S model integrated with \textit{Pioneer Student}.} While maintaining the standard T--S logic, where the teacher $f_{\theta_t}$ processes weakly augmented image $\alpha(x_u)$ to generate pseudo-label $q_u$ as supervision for the student's prediction $p_s$ on strongly augmented sample $\mathcal{A}(x_u)$, we introduce a parameter-independent PiS model $f_{\theta_{\pi}}$. The PiS accepts extremely augmented image $\mathbb{A}(x_u)$ and aligns its prediction $p_{\pi}$ with the teacher's supervision. Periodically, the parameters of the T--S model $\theta_s$ are overwritten by the parameters of the pioneer student $\theta_{\pi}$.}
    \label{fig:pis}
      \vspace{-8pt}
\end{figure*}

Both FGD and LDT appear to share a common underlying issue: \textit{a single parameter space} is required to simultaneously satisfy \textit{two conflicting optimization demands}, resulting in optimization coupling.
To address this, we augment the vanilla T--S framework with an additional, parameter-independent branch: the Pioneer Student $f_{\theta_{\pi}}$. 
As illustrated in Figure~\ref{fig:pis}, PiS is trained under teacher supervision and periodically teleports its accumulated knowledge back to the T--S model via parameter overwrite.

\subsection{Independent Parameter Space for Decoupled Optimization}

\textbf{Pioneer Student Branch.} Since the parameters $\theta_{\pi}$ of PiS are isolated and not instantaneously coupled with the T--S model during training, this design creates room for adopting aggressive generalization strategies. Specifically, we employ extreme augmentation $\mathbb{A}$ (see Section~\ref{sec:sens_ana} for definitions) to promote robustness. 
Given an unlabeled sample $x_u$, PiS takes its extremely augmented version $\mathbb{A}(x_u)$ as input to generate the prediction $p_{\pi} = f_{\theta_{\pi}}(\mathbb{A}(x_u))$. The optimization objective is defined as:
\begin{equation}
    \mathcal{L}_{\pi}(x_u, q_u; \theta_{\pi}) = \mathcal{H}(q_u, \sigma(p_{\pi})),
\end{equation}
where the pseudo-label supervision $q_u$ is still provided by the teacher network $f_{\theta_t}$. Crucially, the parameter update process of PiS is \emph{exclusively} driven by the unlabeled loss $\mathcal{L}_{\pi}$:
\begin{equation}
    \theta_{\pi} \leftarrow \theta_{\pi} - \eta \nabla_{\theta_{\pi}} \mathcal{L}_{\pi}.
\end{equation}
Through this design, PiS simultaneously resolves both optimization couplings. Since $\theta_{\pi}$ is not instantaneously coupled with the teacher parameters, the PiS’s generalization optimization does not continuously compromise the teacher’s fitting fidelity, thereby alleviating the FGD.
Meanwhile, since the optimization of $\theta_{\pi}$ is driven exclusively by $\mathcal{L}_{\pi}$, it is free from interference from non-global optimization signals induced by labeled loss, thereby mitigating the LDT.

\subsection{Periodic Knowledge Transfer}
\label{sec:knowledge}

\textbf{Periodic Parameter Overwrite.} At first glance, PiS appears capable of resolving both issues on its own. However, it remains under the supervision of the teacher. If left in a purely passive state of supervision, the inherent biases of the teacher would gradually propagate to PiS, eventually dragging its performance toward mediocrity. To address this, we propose periodic parameter overwrite. Specifically, the training procedure is partitioned into multiple stages, each spanning a fixed proportion $\rho$ of the total training progress. At the end of each stage, the PiS transfers its accumulated robust knowledge back to the T--S model:
\begin{equation}
    \quad \theta_s \leftarrow \theta_{\pi}.
\end{equation}
By teleporting this robust knowledge back into the T--S framework, we expect to stage-wise raise the performance ceiling of the T--S model. This, in turn, leads to higher-quality pseudo-labels for PiS in subsequent stages, forming a positive feedback loop.

\textbf{Pioneer Student Warm-Up.} Since PiS is tasked with injecting its knowledge back into the T--S model, the reliability of its knowledge is important. Predictably, the convergence of PiS is relatively slow during the initial training phase because: 1) its optimization is entirely detached from the explicit guidance of ground-truth labels; and 2) the pseudo-labels $q_u$ generated by the teacher at early stages are of low quality. Enforcing parameter overwrite during this period could lead to a collapse in the T--S model's performance. Therefore, we introduce a warm-up mechanism: during the initial $\phi$ fraction of the total training progress, the parameter overwrite is disabled. This ensures that PiS intervenes in the T--S model's optimization only after establishing sufficiently reliable representations. The parameter overwrite schedule with warm-up is formalized as:
\begin{equation}
\theta_s \leftarrow \theta_{\pi} \quad \text{if} \quad \tau = \phi + n\rho, \quad n = 1, 2, \dots, \left\lfloor \frac{1-\phi}{\rho} \right\rfloor,
\end{equation}
where $\tau \in [0, 1]$ represents the current training progress, and $n$ denotes the index of the overwrite stage following the warm-up phase.

\textbf{Parameter Update Sharing.} 
Although the warm-up strategy addresses the unreliability of early-stage knowledge, in rare SSL scenarios where labels are highly abundant, PiS that relies solely on unlabeled data may fail to match the T--S model that fully exploits the rich supervision throughout training.
To ensure the robustness and applicability of PiS across diverse regimes, we propose parameter update sharing, which leverages the architectural consistency between the T--S student and PiS. Specifically, in label-abundant settings, we record the gradient vector of the student network generated by the joint optimization loss (i.e., Equation~\ref{equ:joint}) at each iteration. This gradient is then treated as an auxiliary update signal and weighted into the PiS's own gradient derived from the unlabeled loss:
\begin{equation}
    \theta_{\pi} \leftarrow \theta_{\pi} - \eta \left( \nabla_{\theta_{\pi}} \mathcal{L}_{\pi} + \beta \nabla_{\theta_s} (\mathcal{L}_l + \mathcal{L}_u) \right),
\end{equation}
where $\beta$ is a hyperparameter controlling the intensity of the update sharing. By reusing the student's already computed gradient, PiS avoids redundant computation on labeled data, minimizing extra overhead.
Meanwhile, the gradient from the optimization-leading student provides a useful optimization guidance for PiS.
As shown in Figure~\ref{fig:UC}, when labeled data is abundant, the UC gap between labeled and unlabeled losses is small, making it difficult for the labeled loss to dominate and form local basin. Even if such a basin emerge, \textit{abundant} labeled data ensures they are more beneficial than detrimental to global optimization.
Consequently, sharing this gradient allows PiS to keep pace with the T--S model without reintroducing the LDT.

\section{Experiments}
\label{sec:Experiments}

All experiments are conducted on the Unified Semi-supervised Learning Benchmark (USB)~\cite{wang2022usb}. The experimental settings and training backbone~\cite{vaswani2017attention} follow prior work~\cite{han2025regmixmatch}, including teacher synchronization via direct duplication ($\theta_t \leftarrow \theta_s$). For the hyperparameters of PiS, we set the stage proportion (i.e., the parameter overwrite interval) to $\rho = 2.5\%$ and the warm-up fraction to $\phi = 8.0\%$ of the total training progress across all tasks. The update sharing intensity is fixed to $\beta = 0.6$ when activated. \textbf{Unless explicitly specified, update sharing is disabled by default for all other experiments}. To ensure reproducibility, each algorithm is executed three times with different random seeds. Detailed configurations are provided in Appendix~\ref{appendix:training_configs}.

\begin{table*}[t]
	\centering
    \caption{\textbf{Top-1 accuracy (\%) across CV, audio, and NLP benchmarks.} We report the mean and standard deviation over three runs. ${\color{green}\uparrow}$ indicates performance improvement after integrating PiS. \textbf{Bold} and \underline{underline} denote the best and second-best results, respectively. $^\ast$ denotes our re-implementation. RegMixMatch results are omitted for audio and NLP due to Mixup incompatibility in these modalities.}
	\label{tab:pis+ssl}
	\resizebox{\textwidth}{!}{%
		\begin{tabular}{l|cc|cc|cc|cc|cc}
			\toprule
			Modality
			& \multicolumn{6}{c|}{CV}
			& \multicolumn{2}{c|}{Audio}
			& \multicolumn{2}{c}{NLP} \\

            \cmidrule(lr){1-1}
			\cmidrule(lr){2-7}
			\cmidrule(lr){8-9}
			\cmidrule(lr){10-11}
			
			Dataset 
			& \multicolumn{2}{c|}{CIFAR-100}
			& \multicolumn{2}{c|}{STL-10}
			& \multicolumn{2}{c|}{EuroSAT}
			& \multicolumn{2}{c|}{UrbanSound8K}
			& \multicolumn{2}{c}{ACL-IMDb} \\

            \cmidrule(lr){1-1}
			\cmidrule(lr){2-3}
			\cmidrule(lr){4-5}
			\cmidrule(lr){6-7}
			\cmidrule(lr){8-9}
			\cmidrule(lr){10-11}
			
			\# Labels per Class 
			& 2 & 4 
			& 4 & 10 
			& 2 & 4 
			& 10 & 40
			& 10 & 50 \\
			\midrule
			
			\multicolumn{11}{l}{\textit{\textbf{Baseline Methods}}} \\
			
			FlexMatch \cite{zhang2021flexmatch} 
			& 73.24{\scriptsize $\pm$1.12} & 81.76{\scriptsize $\pm$0.36}
			& 85.60{\scriptsize $\pm$3.11} & 91.83{\scriptsize $\pm$0.78}
			& 94.83{\scriptsize $\pm$0.57} & 94.42{\scriptsize $\pm$0.81}
			& 60.77{\scriptsize $\pm$0.96} & 76.30{\scriptsize $\pm$1.66}
			& 92.18{\scriptsize $\pm$0.77} & 92.59{\scriptsize $\pm$0.38} \\
			
			CoMatch \cite{li2021comatch} 
			& 64.92{\scriptsize $\pm$0.69} & 74.77{\scriptsize $\pm$0.50}
			& 84.88{\scriptsize $\pm$1.88} & 90.44{\scriptsize $\pm$1.35}
			& 94.25{\scriptsize $\pm$0.43} & 95.19{\scriptsize $\pm$1.05}
			& \textbf{70.37}{\scriptsize $\pm$1.15} & 79.81{\scriptsize $\pm$2.34}
			& \underline{92.56}{\scriptsize $\pm$0.30} & 92.38{\scriptsize $\pm$1.14} \\
			
			SimMatch \cite{zheng2022simmatch} 
			& 76.22{\scriptsize $\pm$1.08} & 82.94{\scriptsize $\pm$0.78}
			& 88.23{\scriptsize $\pm$3.20} & 92.45{\scriptsize $\pm$1.86}
			& 92.34{\scriptsize $\pm$0.60} & 94.73{\scriptsize $\pm$0.89}
			& 63.37{\scriptsize $\pm$6.10} & \textbf{80.05}{\scriptsize $\pm$2.11}
			& 92.07{\scriptsize $\pm$0.55} & \underline{92.92}{\scriptsize $\pm$0.33} \\
			
			SoftMatch \cite{chen2023softmatch} 
			& 77.33{\scriptsize $\pm$1.32} & 83.16{\scriptsize $\pm$0.66}
			& 86.45{\scriptsize $\pm$3.16} & 92.16{\scriptsize $\pm$1.72}
			& 94.25{\scriptsize $\pm$0.62} & 94.10{\scriptsize $\pm$1.42}
			& 64.68{\scriptsize $\pm$6.59} & 78.57{\scriptsize $\pm$2.37}
			& 92.24{\scriptsize $\pm$0.58} & 92.03{\scriptsize $\pm$0.72} \\
			
			SequenceMatch$^\ast$ \cite{nguyen2024sequencematch} 
			& 78.52{\scriptsize $\pm$0.46} & 83.69{\scriptsize $\pm$0.23}
			& 87.77{\scriptsize $\pm$2.61} & 92.16{\scriptsize $\pm$0.55}
			& 94.12{\scriptsize $\pm$0.72} & 94.99{\scriptsize $\pm$0.69}
			& - & -
			& - & - \\
			
			\midrule
			\multicolumn{11}{l}{\textit{\textbf{Plug-in Evaluation (Ours)}}} \\
			
			FixMatch$^\ast$ \cite{sohn2020fixmatch} 
			& 70.39{\scriptsize $\pm$0.88} & 80.16{\scriptsize $\pm$0.71}
			& 83.89{\scriptsize $\pm$2.06} & 91.89{\scriptsize $\pm$0.70}
			& 86.69{\scriptsize $\pm$4.05} & 94.00{\scriptsize $\pm$2.27}
			& 60.08{\scriptsize $\pm$5.72} & 79.16{\scriptsize $\pm$2.13}
			& 92.26{\scriptsize $\pm$0.27} & 92.65{\scriptsize $\pm$0.13} \\
			
			\rowcolor{lightgray!50}\quad + PiS 
			& 73.99{\scriptsize $\pm$0.94}$^{\color{green}\uparrow}$  
			& 82.02{\scriptsize $\pm$0.37}$^{\color{green}\uparrow}$  
			& 86.83{\scriptsize $\pm$3.49}$^{\color{green}\uparrow}$  
			& \textbf{93.65}{\scriptsize $\pm$0.51}$^{\color{green}\uparrow}$  
			& 88.71{\scriptsize $\pm$1.62}$^{\color{green}\uparrow}$  
			& 95.04{\scriptsize $\pm$0.15}$^{\color{green}\uparrow}$  
			& 66.87{\scriptsize $\pm$4.88}$^{\color{green}\uparrow}$  
			& \underline{79.95}{\scriptsize $\pm$2.56}$^{\color{green}\uparrow}$  
			& \textbf{92.59}{\scriptsize $\pm$0.30}$^{\color{green}\uparrow}$  
			& \textbf{92.98}{\scriptsize $\pm$0.22}$^{\color{green}\uparrow}$  \\

			FreeMatch$^\ast$ \cite{wang2023freematch} 
			& 78.46{\scriptsize $\pm$0.77} & 83.58{\scriptsize $\pm$0.63}
			& 87.20{\scriptsize $\pm$2.96} & 91.51{\scriptsize $\pm$0.33}
			& 93.48{\scriptsize $\pm$0.78} & 94.20{\scriptsize $\pm$0.55}
			& 60.23{\scriptsize $\pm$4.56} & 75.85{\scriptsize $\pm$0.78}
			& 91.31{\scriptsize $\pm$0.25} & 92.09{\scriptsize $\pm$0.57} \\
			
			\rowcolor{lightgray!50}\quad + PiS 
			& \underline{80.83}{\scriptsize $\pm$1.64}$^{\color{green}\uparrow}$  
			& \textbf{84.45}{\scriptsize $\pm$0.44}$^{\color{green}\uparrow}$  
			& \underline{90.65}{\scriptsize $\pm$0.22}$^{\color{green}\uparrow}$  
			& 93.13{\scriptsize $\pm$0.15}$^{\color{green}\uparrow}$  
			& 94.92{\scriptsize $\pm$1.45}$^{\color{green}\uparrow}$  
			& 95.89{\scriptsize $\pm$1.07}$^{\color{green}\uparrow}$  
			& \underline{66.89}{\scriptsize $\pm$3.69}$^{\color{green}\uparrow}$  
			& 79.40{\scriptsize $\pm$1.87}$^{\color{green}\uparrow}$  
			& 92.22{\scriptsize $\pm$0.67}$^{\color{green}\uparrow}$  
			& 92.60{\scriptsize $\pm$0.65}$^{\color{green}\uparrow}$  \\

			RegMixMatch$^\ast$ \cite{han2025regmixmatch} 
			& 80.49{\scriptsize $\pm$0.62} & 83.89{\scriptsize $\pm$0.49}
			& 89.97{\scriptsize $\pm$3.08} & 92.74{\scriptsize $\pm$0.75}
			& \underline{95.62}{\scriptsize $\pm$0.61} & \textbf{96.38}{\scriptsize $\pm$0.74}
			& - & -
			& - & - \\
			
			\rowcolor{lightgray!50}\quad + PiS 
			& \textbf{81.44}{\scriptsize $\pm$0.97}$^{\color{green}\uparrow}$  
			& \underline{84.19}{\scriptsize $\pm$0.43}$^{\color{green}\uparrow}$  
			& \textbf{91.33}{\scriptsize $\pm$1.57}$^{\color{green}\uparrow}$  
			& \underline{93.62}{\scriptsize $\pm$1.10}$^{\color{green}\uparrow}$  
			& \textbf{96.02}{\scriptsize $\pm$0.20}$^{\color{green}\uparrow}$  
			& \underline{96.16}{\scriptsize $\pm$0.60}  
			& - & -
			& - & - \\
			
			\bottomrule
		\end{tabular}
	}
      \vspace{-8pt}
\end{table*}

\subsection{SSL with Pioneer Student}
\label{sec:SSL+pis}

We first present the performance of eight recent SSL methods across three modalities on various datasets: CV (CIFAR-100~\cite{krizhevsky2009learning}, STL-10~\cite{coates2011analysis}, EuroSAT~\cite{helber2019eurosat}), audio (UrbanSound8K~\cite{salamon2014dataset}), and NLP (ACL-IMDb~\cite{maas2011learning}) under varying labeled sample counts. To verify the universality of PiS, we integrate it into three representative baselines: FixMatch~\cite{sohn2020fixmatch}, FreeMatch~\cite{wang2023freematch}, and the state-of-the-art RegMixMatch~\cite{han2025regmixmatch}.

As shown in Table~\ref{tab:pis+ssl}, PiS consistently improves performance across diverse settings. In the CV domain, PiS brings notable gains in low-label regimes. For FixMatch, it achieves +3.60\% on CIFAR-100 (2 labels) and +2.94\% on STL-10 (4 labels), while reaching the best result on STL-10 (93.65\%). Similarly, FreeMatch is consistently improved by PiS, including +2.37\% on CIFAR-100 (2 labels) and +3.45\% on STL-10 (4 labels), achieving competitive or state-of-the-art performance. Importantly, PiS also enhances the strong RegMixMatch baseline, yielding the best or second-best results across most CV settings. Beyond CV, PiS generalizes effectively to other modalities. It significantly improves performance on UrbanSound8K for FixMatch and FreeMatch (+6.79\% and +6.66\% at 10 labels), and further refines performance on the near-saturated ACL-IMDb benchmark. Overall, these results demonstrate that PiS is a simple yet broadly applicable plug-in for SSL.

\begin{wrapfigure}{r}{0.5\textwidth} 
  \vspace{-14pt} 
  \centering
  \captionsetup[subfigure]{font=scriptsize,labelfont=scriptsize}
  
  \begin{subfigure}{0.48\linewidth}
    \centering
    \includegraphics[width=\linewidth]{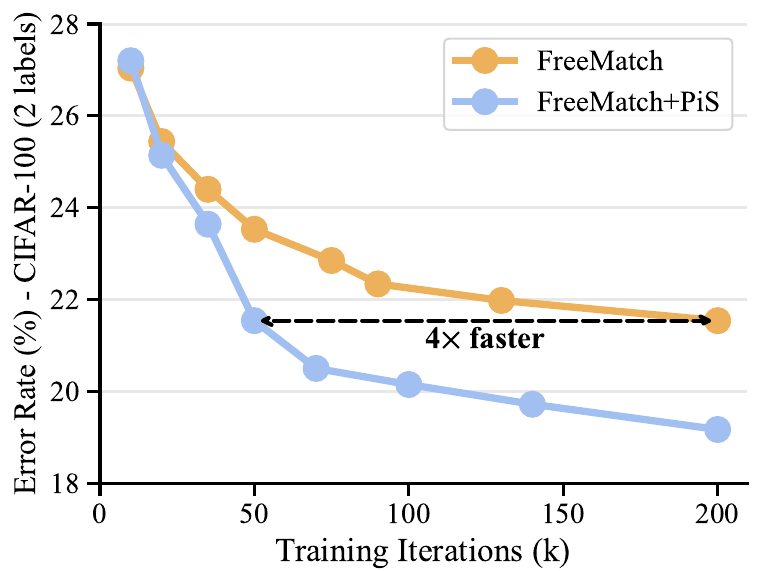}
  \end{subfigure}
  \hfill
  \begin{subfigure}{0.48\linewidth}
    \centering
    \includegraphics[width=\linewidth]{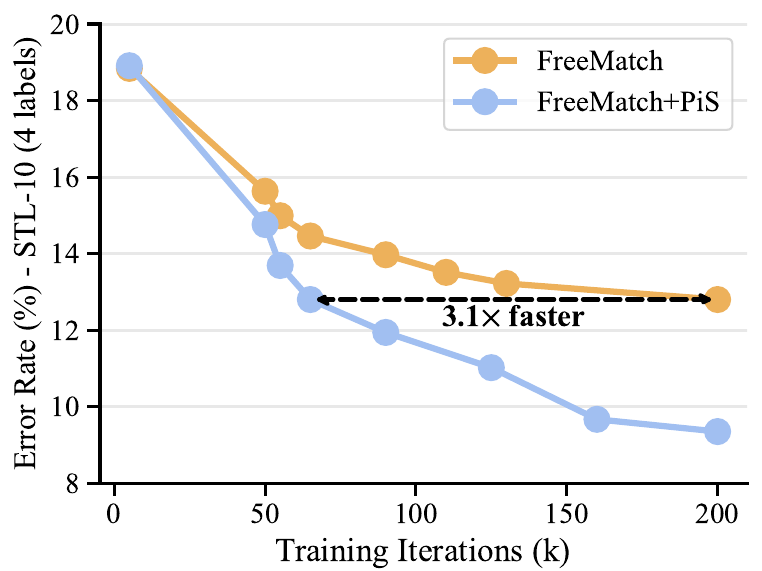}
  \end{subfigure}
  
  \vspace{5pt}
  \caption{\textbf{PiS accelerates convergence by reaching the same performance with significantly fewer training iterations}.}
  \label{fig:pis_eff}
  \vspace{-10pt}
\end{wrapfigure}


Finally, while PiS introduces non-negligible per-iteration overhead (ranging from 20\% to 36\% depending on the baseline, see Appendix~\ref{appendix:overhead} for details), it significantly accelerates convergence. As shown in Figure~\ref{fig:pis_eff}, FreeMatch+PiS reaches the performance of FreeMatch at 200k iterations within only 50k (CIFAR-100) and 65k (STL-10) iterations, corresponding to a $3.1\times$–$4\times$ speedup. This improved optimization efficiency largely offsets the additional cost, making PiS a practical and cost-effective solution for SSL. Moreover, PiS maintains a clear advantage even under equal training cost (Appendix~\ref{appendix:overhead}, Table~\ref{tab:equal_time}).

\begin{wraptable}{r}{0.34\textwidth} 
    \centering
    \vspace{-12pt} 
	\caption{\textbf{Top-1 accuracy (\%) on ImageNet.} $^\dagger$ denotes the FreePiS w/ parameter update sharing.}
	\label{tab:Imagenet}
    \resizebox{0.34\textwidth}{!}{ 
        \begin{tabular}{l|cc}
			\toprule
			Dataset & \multicolumn{2}{c}{ImageNet} \\
			\cmidrule(lr){1-1} \cmidrule(lr){2-3}
			\# Labels per Class & 10 & 100 \\ \cmidrule{1-3}
			FreeMatch & 54.69 & 72.57 \\
			RegMixMatch & 58.35 & \textbf{73.66}\\ \midrule
			FreePiS & \textbf{60.02} & 66.69\\
			FreePiS$^{\dagger}$ & -- & 73.24\\
			\bottomrule
        \end{tabular}
    }
    \vspace{-8pt} 
\end{wraptable}

\subsection{Applicability Analysis of PiS}
\label{sec:when_excel}

While PiS consistently improves SSL performance across standard benchmarks, we further evaluate its behavior under a broader range of supervision densities to assess its applicability boundaries. Note that all subsequent results are based on integrating PiS into FreeMatch, which we refer to as \textit{FreePiS}, \textbf{and are compared against other FreeMatch-based counterparts.}

\textbf{ImageNet.} Table~\ref{tab:Imagenet} reports results on ImageNet~\cite{deng2009imagenet} under two supervision densities.
Under the 10-label-per-class setting, FreePiS achieves the best result with a substantial +5.33\% gain over FreeMatch, showing its effectiveness in the low-label regime where LDT is most pronounced.
Under the label-abundant setting (100 labels/class, totaling 100,000 labeled data), however, FreePiS exhibits a noticeable performance drop.
This is consistent with our analysis in Section~\ref{sec:knowledge}: when labeled data is plentiful, a PiS branch that exclusively optimizes the unlabeled loss discards highly reliable supervision, causing it to converge slower and subsequently inject inferior knowledge into the T--S model.
This failure case directly motivates the parameter update sharing mechanism.
When parameter update sharing is enabled (FreePiS$^{\dagger}$), PiS is guided by the more informative joint-loss gradient of the T--S student, allowing it to keep pace and serve as a reliable knowledge source.
As a result, FreePiS$^{\dagger}$ recovers to 73.24\%, effectively improving upon the baseline.
The remaining gap relative to RegMixMatch is expected: in the label-abundant regime, the LDT effect is inherently weak, making the mitigation of FGD the primary advantage of PiS rather than the dual benefit in other settings.

\begin{figure}[t]
  \centering
  \captionsetup[subfigure]{font=scriptsize,labelfont=scriptsize}
  
  \begin{subfigure}{0.24\linewidth}
    \includegraphics[width=\linewidth]{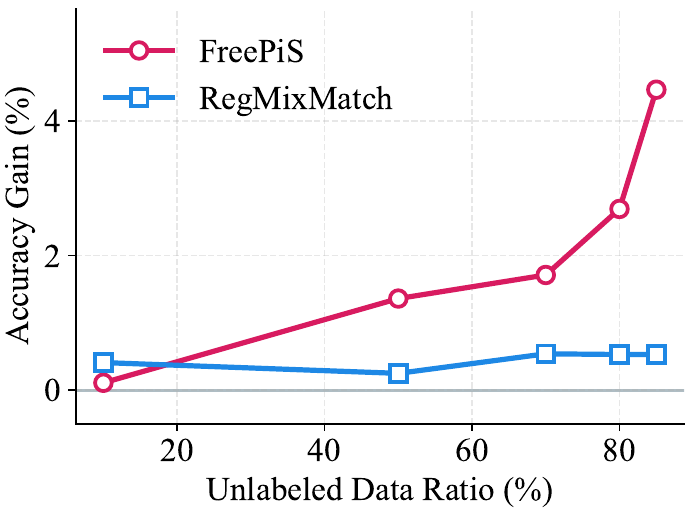}
    \caption{Varying Unlabeled Data.} \label{fig:lb10}
  \end{subfigure}
  \begin{subfigure}{0.24\linewidth}
    \includegraphics[width=\linewidth]{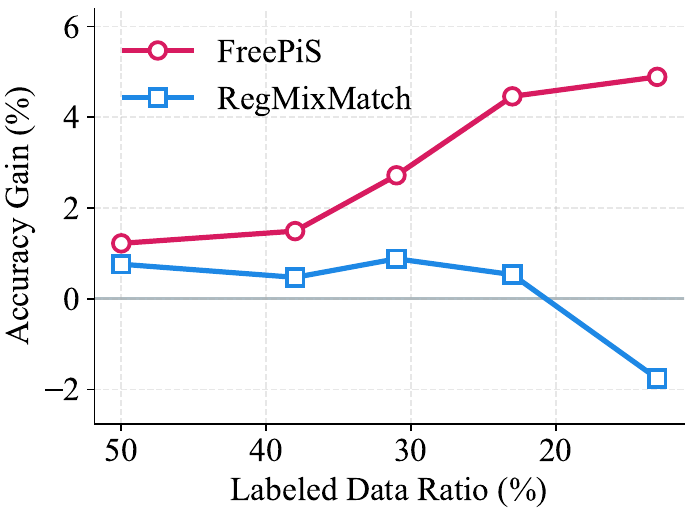}
    \caption{Varying Labeled Data.} \label{fig:ulb66}
  \end{subfigure}
  \hfill 
  \begin{subfigure}{0.24\linewidth}
    \includegraphics[width=\linewidth]{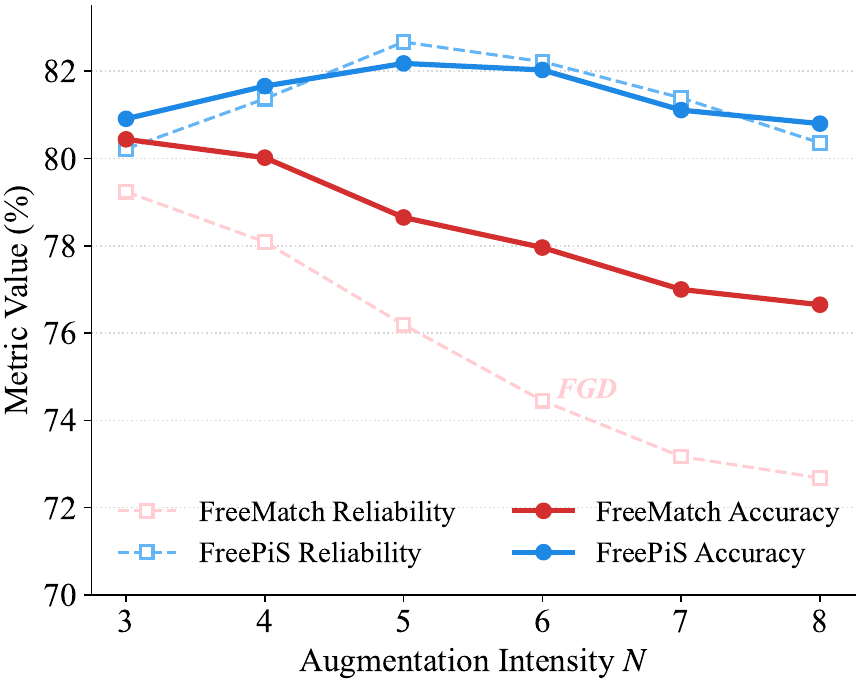}
    \caption{Augmentation Intensity.} \label{fig:aug_fgd}
  \end{subfigure}
  \begin{subfigure}{0.24\linewidth}
    \includegraphics[width=\linewidth]{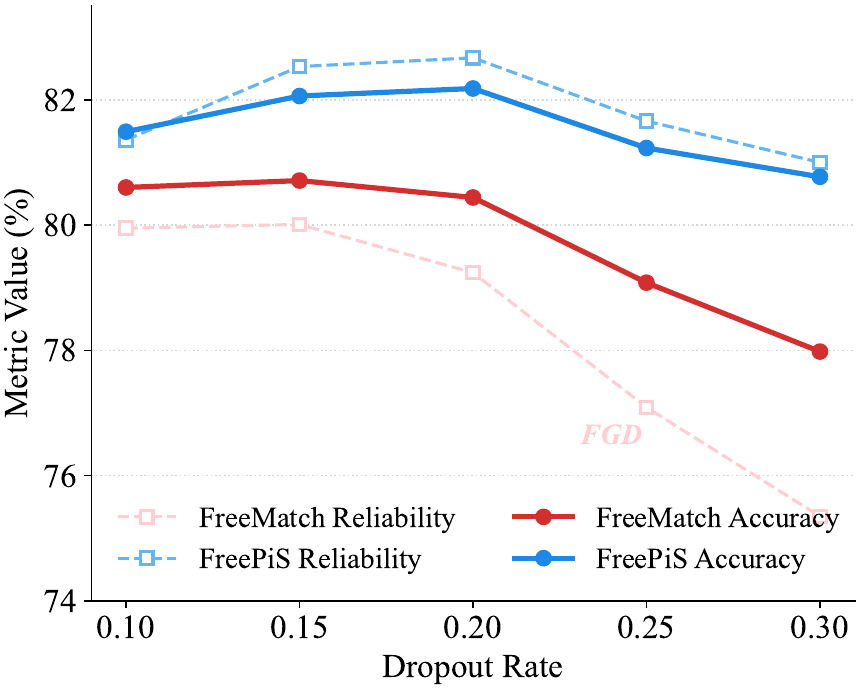}
    \caption{Dropout Rate.} \label{fig:dp_fgd}
  \end{subfigure}

\caption{\textbf{Comparative study of model performance under different data scales and regularization intensities.} (a--b) Performance gains across varying scales of labeled and unlabeled data. (c--d) Teacher's pseudo-label reliability and final accuracy under different augmentation intensities and dropout rates~\cite{srivastava2014dropout} (see Appendix~\ref{app:tab_fgd} for numerical results; see Section~\ref{sec:sens_ana} for augmentation details).}
  \label{fig:combined}
  \vspace{-5pt}
\end{figure}

\textbf{Varying Labeled and Unlabeled Data.} To further assess the applicability of PiS under varying data regimes, we systematically study its performance across different data scales. Specifically, on the FGVC dataset~\cite{fgvc}, we examine two complementary scenarios: 1) fixing a constant budget of labeled data at $N_l = 10$ per class while varying the unlabeled volume (Figure~\ref{fig:lb10}), and 2) fixing the unlabeled data at $N_u = 33$ per class while adjusting the labeled data ratio (Figure~\ref{fig:ulb66}). Parameter update sharing is activated when the labeled data ratio $\geq 30\%$. We report the \textbf{performance gains} of RegMixMatch and FreePiS relative to the FreeMatch baseline.

As illustrated in Figure~\ref{fig:combined}(a--b), the performance gains yielded by FreePiS become increasingly significant as the label ratio decreases or the unlabeled data scale expands. In contrast, despite employing Mixup~\cite{zhang2018mixup}, the gains from RegMixMatch are limited or even negative in certain cases. This finding validates the unique advantage of PiS: \textbf{in real-world SSL scenarios characterized by sparse labels and massive unlabeled data, the PiS demonstrates exceptional practical value.}

\subsection{Dual Benefits of Pioneer Student}
\label{sec:dualb}

\textbf{Overcoming the Fitting--Generalization Dilemma.} 
Revisiting the FGD: the student's pursuit of strong generalization often comes at the expense of the teacher's fitting fidelity. In Figure~\ref{fig:combined}(c--d), we illustrate the changes in teacher's reliability (i.e., pseudo-labels accuracy) and classification accuracy on the CIFAR-100 dataset (2 labels per class), where we increase the generalization intensity for the student in FreeMatch and the PiS branch in FreePiS respectively. It is observed that for FreeMatch, aggressive generalization strategies significantly weaken the teacher's reliability, thereby suppressing overall classification performance. In contrast, integrating the PiS branch allows the framework to accommodate both radical generalization strategies and high-level teacher fitting fidelity, suggesting that \textbf{PiS effectively elevates the achievable generalization upper bound for SSL.}

\textbf{Avoiding the Label-Driven Trap.} 
As shown in the second row of Figure~\ref{fig:losscape}a, contrary to observations from FreeMatch loss landscapes, incorporating the PiS branch moderates the convergence pace of the labeled loss during early training ($\mathcal{L}_l=0.36$), while simultaneously steering the unlabeled loss toward a more favorable optimization state ($\mathcal{L}_u=1.12$) with a flatter loss landscape. This implies that PiS explicitly emphasizes the role of the unlabeled loss in the global optimization process particularly in the early stages, thereby mitigating the interference of the labeled loss and \textbf{endowing the model with greater optimization potential.} Consequently, as illustrated in the second row of Figure~\ref{fig:losscape}b, the unlabeled loss landscape in the subsequent optimization stages exhibits better flatness ($\mathcal{S} \in [2.6, 2.9]$) compared to FreeMatch, providing direct evidence of its greater generalization.

\begin{table*}[t]
\centering
\caption{\textbf{Ablation study on disentangling the dual advantages of FreePiS.} We evaluate two primary mechanisms: addressing the FGD (Anti-FGD) and overcoming the LDT (Anti-LDT). \cmark~indicates the mechanism is \textbf{activated}. $\Delta$ represents the performance gain relative to the baseline (a). Experiments are conducted on FGVC with 5 labels per class and CIFAR-100 with 2 labels per class.}
\label{tab:ablation_main}
\renewcommand{\arraystretch}{1.00}
\resizebox{0.95\textwidth}{!}{ 
\begin{tabular}{@{}cl ccc cc cc@{}}
\toprule
\multirow{2}{*}{ID} & \multirow{2}{*}{Method/Variant} & \multicolumn{3}{c}{Mechanisms} & \multicolumn{2}{c}{FGVC} & \multicolumn{2}{c}{CIFAR-100} \\ \cmidrule(lr){3-5} \cmidrule(lr){6-7} \cmidrule(lr){8-9} 
 & & Anti-FGD & Anti-LDT & Warm-up & Acc (\%) & $\Delta$ & Acc (\%) & $\Delta$ \\ \midrule
(\textit{a}) & FreeMatch & \xmark & \xmark & -- & 51.65 & -- & 78.46 & -- \\ \midrule
(\textit{b}) & FreePiS w. Label Loss & \cmark & \xmark & \cmark & 54.66 & +3.01 & 79.02 & +0.56 \\
(\textit{c}) & FreePiS w. Same Aug. & \xmark & \cmark & \cmark & 52.78 & +1.13 & 80.57 & +2.11 \\
(\textit{d}) & FreePiS w/o Warm-up & \cmark & \cmark & \xmark & 52.25 & +0.60 & 80.66 & +2.20 \\ \midrule
\rowcolor{gray!8}
(\textit{e}) & \textbf{FreePiS (Full)} & \cmark & \cmark & \cmark & \textbf{55.54} & \textbf{+3.89} & \textbf{80.83} & \textbf{+2.37} \\ \bottomrule
\end{tabular}
}
\vspace{-5pt}
\end{table*}

\textbf{Disentangling the Dual Advantages.} 
Since PiS inherently addresses both issues, we design two variants to isolate their individual effects: 1) \textit{FreePiS w. Label Loss}: The PiS is optimized with labeled loss in addition to teacher supervision (targeting only the FGD); 2) \textit{FreePiS w. Same Aug.}: The PiS adopts the same augmentation intensity as the student (targeting only the LDT). As reported in Table~\ref{tab:ablation_main}, the primary source of improvement varies across datasets. On FGVC, mitigating the FGD (comparing \textit{a} and \textit{b}) accounts for the majority of the performance gain (+3.01\%). In contrast, on CIFAR-100, addressing the LDT (comparing \textit{a} and \textit{c}) provides a more significant contribution (+2.11\%). Notably, comparing variants \textit{b} and \textit{e} reveals an interesting yet consistent finding: incorporating additional labeled loss into the PiS optimization leads to a 1.81\% performance degradation on CIFAR-100. This finding offers direct empirical evidence of \textbf{the adverse impact that labeled data can exert on the SSL optimization.} Finally, comparing variants \textit{d} and \textit{e} highlights the necessity of a warm-up phase, with its removal leading to a substantial performance drop on FGVC (-3.29\%).

\subsection{Hyperparameter Sensitivity Analysis}
\label{sec:sens_ana}

\begin{wrapfigure}{r}{0.48\textwidth} 
  \vspace{-5pt} 
  \centering
  \captionsetup[subfigure]{font=scriptsize,labelfont=scriptsize}
  
  \begin{subfigure}{0.48\linewidth}
    \centering
    \includegraphics[width=\linewidth]{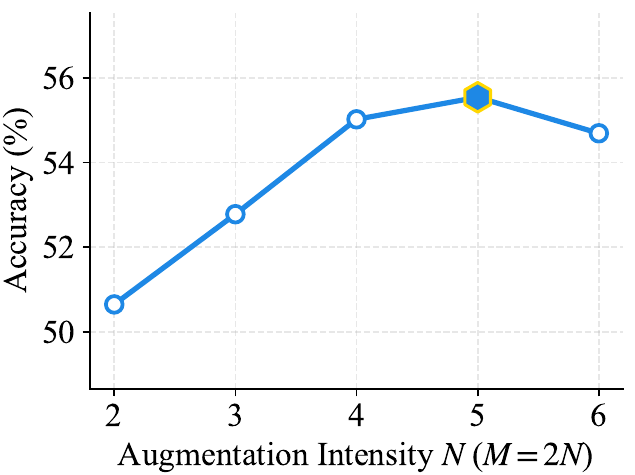}
    \caption{Augmentation Intensity.} \label{fig:aug_abla}
  \end{subfigure}
  \hfill
  \begin{subfigure}{0.48\linewidth}
    \centering
    \includegraphics[width=\linewidth]{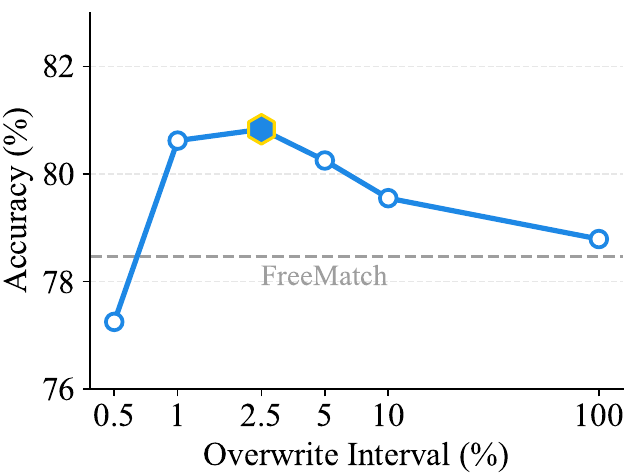}
    \caption{Overwrite Interval.} \label{fig:ite_abla}
  \end{subfigure}
  
  \vspace{5pt}
  \caption{\textbf{Sensitivity analysis of FreePiS hyperparameters.} (a) PiS's data augmentation intensity $N$ on FGVC (5 labels). (b) Parameter overwrite interval on CIFAR-100 (2 labels).}
  \label{fig:abla}
  \vspace{-10pt}
\end{wrapfigure}

\textbf{Augmentation Intensity.} 
SSL typically employs RandAugment~\cite{cubuk2020randaugment} as the data augmentation strategy, with parameters $N$ and $M$ controlling the intensity. To simplify hyperparameter tuning, we fix $M=2N$ and vary the overall intensity solely by adjusting $N$. Figure~\ref{fig:aug_abla} depicts the classification accuracy versus augmentation intensity on the FGVC dataset; the model reaches its peak performance at $N=5$. Accordingly, we apply \texttt{RandAugment(5, 10)} for the PiS branch across all tasks.\footnote{Following standard SSL practice, we use \texttt{RandAugment(3, 5)} as the default strong augmentation.}

\textbf{Parameter Overwrite Interval.} 
Periodic parameter overwrite requires the PiS to transfer its parameters to the teacher at specific training intervals. 
Intuitively, overly frequent overwriting may impair the teacher’s fitting capacity and destabilize training, 
whereas an excessively long interval causes the knowledge acquired by the PiS to be progressively eroded by teacher supervision. As shown in Figure~\ref{fig:ite_abla}, omitting periodic overwrite (i.e., setting the interval to 100\%) results in performance close to the FreeMatch baseline (dashed line) and\textbf{ substantially lower than the peak achieved at the optimal interval} (2.5\%). These results suggest that parameter overwrite stage-wise raises the performance ceiling of the T--S model, leading to further overall improvement. Finally, although parameter overwrite is conceptually simple, we show in Appendix~\ref{appendix:sync} that replacing it with softer synchronization mechanisms consistently underperforms direct overwrite, underscoring that the hard overwrite is essential for forcefully teleporting the model out of labeled-dominated local minima.

\section{Conclusion}
\label{sec:conclusion}
This paper identifies two optimization pathologies in conventional SSL, the FGD and the LDT, that arise from forcing conflicting optimization demands into a single parameter space. The proposed PiS addresses both issues by operating in an independent parameter space to achieve decoupled optimization, and by periodically injecting its learned knowledge into the T--S framework. Empirically, PiS consistently improves mainstream SSL methods as a plug-and-play module, with particularly pronounced gains in practical SSL scenarios characterized by sparse labels and abundant unlabeled data. We hope this work
provides useful insights for the architectural design of SSL systems.

\textbf{Limitations and Future Directions.}
As an auxiliary branch, PiS introduces non-negligible computational overhead during training, motivating future work on more cost-effective decoupled designs. Furthermore, we hope the concept of optimization decoupling can be extended to other SSL tasks (e.g., dense prediction) and even broader paradigms beyond SSL that utilize T--S architectures.


{
\small
\bibliographystyle{unsrtnat} 
\bibliography{neurips}          
}



\appendix

\section{Gradient Magnitude Analysis Supporting the LDT}
\label{appendix:grad_norm}

The UC metric (Eq.~\ref{equ:uc}) characterizes only the directional coherence of gradient updates and is invariant to gradient magnitude. A natural question therefore arises: although the labeled gradients exhibit substantially higher UC, could their magnitudes be negligible compared to the unlabeled gradients, thereby weakening their influence on the optimization trajectory?

\begin{wrapfigure}{r}{0.48\textwidth} 
  \vspace{-12pt} 
  \centering
  \captionsetup[subfigure]{font=scriptsize,labelfont=scriptsize}
  
  \begin{subfigure}{0.48\linewidth}
    \centering
    \includegraphics[width=\linewidth]{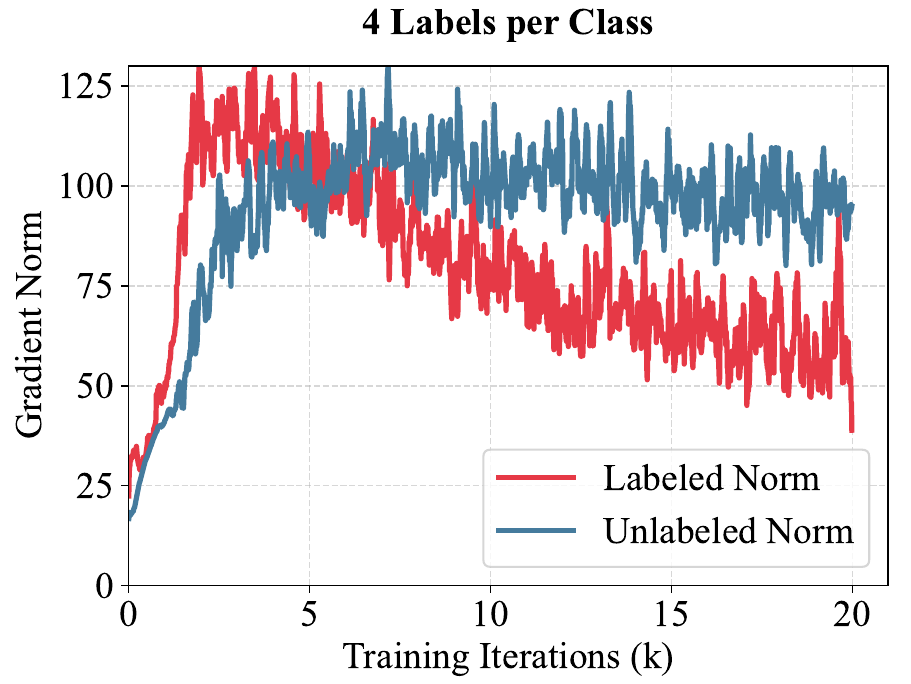}
  \end{subfigure}
  \hfill
  \begin{subfigure}{0.48\linewidth}
    \centering
    \includegraphics[width=\linewidth]{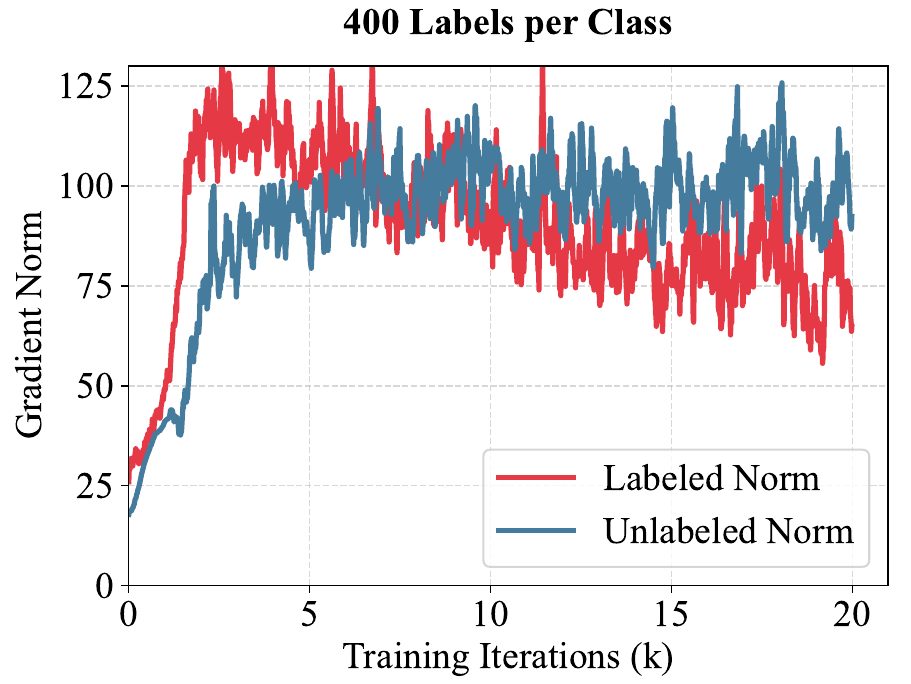}
  \end{subfigure}
  
  \caption{Gradient norms of labeled and unlabeled losses.}
  \label{fig:grad_norm}
  \vspace{-14pt}
\end{wrapfigure}

To address this concern, we visualize the evolution of the gradient norms
$\|\nabla_{\theta_s}\mathcal{L}_l\|$
and
$\|\nabla_{\theta_s}\mathcal{L}_u\|$
during the early training stage of FreeMatch on STL-10 under both the 4-label/class and 400-label/class settings (Figure~\ref{fig:grad_norm}).

We observe that, in both settings, the labeled gradient norm is slightly larger than the unlabeled counterpart at the beginning of training, and remains broadly comparable throughout the early optimization stage before the unlabeled gradient gradually becomes dominant later in training.

These observations support two important conclusions.
First, the labeled gradients are not negligible in magnitude during the early stage; they contribute optimization steps of comparable scale to the unlabeled gradients. This validates that the labeled objective exerts a substantial influence on the shared parameter updates.

Second, because the gradient norms are broadly comparable during the critical early stage, the primary asymmetry between the two optimization signals lies in their directional behaviors. In this regime, optimization dynamics are increasingly governed by directional consistency rather than instantaneous magnitude. Consequently, the significantly higher UC of the labeled gradients explains why the shared parameters are preferentially steered toward a labeled-dominated local basin, giving rise to the Label-Driven Trap.

\section{Quantifying Loss Sharpness and Generalization}
\label{appendix:sharpness}

\subsection{Loss Flatness and Generalization}
The geometry of the loss landscape is closely related to the generalization capability of deep learning models~\cite{keskar2016large,neyshabur2017exploring,li2018visualizing}. A widely accepted hypothesis in optimization literature is that ``flat'' local minima generalize better than ``sharp'' ones. In a flat region, the loss value remains low even when model parameters $\boldsymbol{w}$ are subject to small perturbations $\boldsymbol{\delta}$, indicating robustness to shifts between training and testing data distributions. Conversely, a sharp minimum is characterized by a rapid increase in loss upon slight parameter changes, making it highly sensitive to noise and prone to poor generalization on unseen data.

\subsection{Directional Sharpness Definition}

Following previous work~\cite{li2018visualizing}, we measure sharpness by exploring the loss surface along random directions in a reduced-dimensional subspace. Given a converged parameter point $\boldsymbol{w}$, we select two random orthogonal directions $\boldsymbol{d}_1, \boldsymbol{d}_2 \in \mathbb{R}^n$. To ensure the perturbation is scale-relevant, these directions are normalized to match the global Frobenius norm of the parameter vector $\boldsymbol{w}$. This ensures that the displacement in the parameter space is proportional to the overall magnitude of the learned weights.

We define a local perturbation $\boldsymbol{\delta}(\alpha, \beta) = \alpha \boldsymbol{d}_1 + \beta \boldsymbol{d}_2$, where $\alpha$ and $\beta$ are scalar coefficients. The sharpness $\mathcal{S}$ is then quantified by the maximum loss increase within a local neighborhood $\mathcal{C}_{\epsilon}$:
\begin{equation}
    \mathcal{S} = \max_{(\alpha, \beta) \in \mathcal{C}_{\epsilon}} \left( f(\boldsymbol{w} + \alpha \boldsymbol{d}_1 + \beta \boldsymbol{d}_2) - f(\boldsymbol{w}) \right)
\end{equation}
where $\mathcal{C}_{\epsilon} = \{ (\alpha, \beta) : -\epsilon \le \alpha, \beta \le \epsilon \}$ defines a square region in the 2D subspace. Note that we omit the scaling denominator $(1+f(\boldsymbol{w}))$ used in earlier work~\cite{keskar2016large} because FreeMatch~\cite{wang2023freematch} and FreePiS are evaluated under identical configurations, making the absolute loss increase $\mathcal{S}$ a more direct and transparent measure of the local sharpness of the loss landscape.

\subsection{Implementation Details and Parameters}
For the results and 3D visualizations presented in Figure~\ref{fig:losscape}, we set the neighborhood scale \textbf{$\epsilon = 1.0$}. The loss values are computed on a grid of size $50 \times 50$ within the range $\alpha, \beta \in [-1.0, 1.0]$. The directions $\boldsymbol{d}_1$ and $\boldsymbol{d}_2$ are sampled from a Gaussian distribution, re-orthogonalized, and then scaled such that $||\boldsymbol{d}_1||_2 = ||\boldsymbol{d}_2||_2 = ||\boldsymbol{w}||_2$.


\section{Detailed Training Configurations and Backbones}
\label{appendix:training_configs}

To ensure reproducibility, we provide the comprehensive experimental settings in Table \ref{tab:usbcfg}. Our training protocols and backbones are strictly aligned with the USB \cite{wang2022usb} and the configurations used in RegMixMatch \cite{han2025regmixmatch}. All experiments in Table~\ref{tab:pis+ssl} are conducted on a single NVIDIA RTX 3090 GPU with 24GB memory. Each training run takes approximately 12--20 hours, depending on the dataset and configuration.

\begin{table}[ht]
\centering
\caption{Hyperparameter configurations and backbones for different datasets.}
\label{tab:usbcfg}
\small 
\begin{tabular}{lccccc}
\toprule
Dataset          & CIFAR-100 & STL-10 & EuroSAT & FGVC-Aircraft & ImageNet-1K \\ \midrule
Image Size                & 32                 & 96              & 32               & 224                    & 224                  \\
Backbone (ViT~\cite{dosovitskiy2020image})            & S-P2-32           & B-P16-96        & S-P2-32         & B-P16-224              & B-P16-224            \\ \midrule
Weight Decay              & 5e-4 & 5e-4 & 5e-4   & 5e-2                           & 5e-2              \\
Learning Rate             & 5e-4               & 1e-4            & 5e-5             & 5e-4                   & 3e-4                 \\
Layer Decay Rate          & 0.5                & 0.95            & 1.0              & 0.5                    & 0.5                  \\ \midrule
Labeled Batch Size        & \multicolumn{4}{c}{8}                                 & 32                   \\
Unlabeled Batch Size      & \multicolumn{4}{c}{8}                                 & 32                   \\ \midrule
LR Scheduler              & \multicolumn{5}{c}{$\eta_k = \eta_0 \cos\left(\frac{7\pi k}{16K}\right)$}              \\
Model EMA Momentum        & \multicolumn{5}{c}{0.0}                                                               \\
Prediction EMA Momentum   & \multicolumn{5}{c}{0.999}                                                             \\ \midrule
Weak Augmentation         & \multicolumn{5}{c}{Random Crop, Random Horizontal Flip}                               \\
Strong Augmentation       & \multicolumn{5}{c}{RandAugment \citep{cubuk2020randaugment}}                         \\ \midrule
Pioneer Student  & \multicolumn{5}{c}{Stage Proportion $\rho = 2.5\%$, Warm-up $\phi = 8\%$,}             \\
Hyperparameters  & \multicolumn{5}{c}{Update Sharing Intensity $\beta = 0.6$ (if activated)}             \\ \bottomrule
\end{tabular}
\end{table}

For NLP tasks, we use the pre-trained BERT-Base~\cite{devlin2019bert} as the backbone. The strong augmentation is Back-Translation~\cite{xie2020unsupervised}. For PiS, which requires extreme augmentation, we increase the sampling temperature from 0.9 (standard strong augmentation) to 1.1. For audio tasks, we adopt HuBERT-Base~\cite{hsu2021hubert} as the backbone. The strong augmentation includes random sub-sample, random gain, random pitch, and random speed. For the extreme augmentation in PiS, we increase the augmentation magnitude range during training. Other details strictly follow the USB benchmark.

\section{Computational Overhead Analysis}
\label{appendix:overhead}

Although PiS substantially reduces computational cost compared to co-training methods~\cite{co-training,ke2019dual,rothenberger2023meta} by reusing teacher supervision and optimizing exclusively on unlabeled data, introducing an auxiliary branch still inevitably incurs non-negligible overhead (see Table~\ref{tab:pis_overhead}). Specifically, due to differences in algorithmic design, integrating PiS into FixMatch, FreeMatch, and RegMixMatch results in approximately 33\%, 36\%, and 20\% additional per-iteration training time, respectively. In terms of memory footprint, the PiS branch increases peak GPU memory consumption by approximately 35\%, 36\%, and 18\% compared to the respective baselines.

\begin{table}[t]
\centering
\caption{Computational overhead of PiS on a single NVIDIA RTX 3090 GPU. 
We report per-iteration training time and peak GPU memory consumption for baseline methods and their PiS-augmented variants.}
\label{tab:pis_overhead}
\begin{tabular}{lcccc}
\toprule
\textbf{Method} & \textbf{Time (s/iter)} & \textbf{+PiS} & \textbf{Memory (MiB)} & \textbf{+PiS} \\
\midrule
FixMatch     & 0.0944 & 0.1256 (+33\%) & 3359 & 4534 (+35\%) \\
FreeMatch    & 0.0960 & 0.1306 (+36\%) & 3341 & 4543 (+36\%) \\
RegMixMatch  & 0.1344 & 0.1613 (+20\%) & 4616 & 5446 (+18\%) \\
\bottomrule
\end{tabular}
\end{table}

As discussed in Section~\ref{sec:SSL+pis}, this per-iteration overhead is largely offset by the accelerated convergence. FreePiS requires significantly fewer iterations to reach the baseline's final performance, achieving a $3.1\times$--$4.0\times$ reduction in training iterations. Accounting for the $36\%$ per-iteration time overhead, the corresponding wall-clock speedup is approximately $2.3\times$--$2.9\times$ in practice. 

Furthermore, while the standard USB benchmark evaluates all methods under a fixed budget of 200k iterations, we also conduct experiments with a reduced iteration budget for FreePiS to match the total wall-clock training time of the FreeMatch baseline. As shown in Table~\ref{tab:equal_time}, even under strict wall-clock time parity (i.e., training FreePiS for 147k iterations in total), FreePiS maintains a substantial performance advantage, confirming its practical efficiency.

\begin{table}[t]
	\centering
    \caption{Performance comparison under equal wall-clock training time. FreePiS is trained for fewer iterations to match the total training time of FreeMatch (200k iterations).}
	\label{tab:equal_time}
	\begin{tabular}{l|cc|cc|cc}
		\toprule
		Dataset 
		& \multicolumn{2}{c|}{CIFAR-100}
		& \multicolumn{2}{c|}{STL-10}
		& \multicolumn{2}{c}{EuroSAT} \\
		
		\cmidrule(lr){1-1}
		\cmidrule(lr){2-3}
		\cmidrule(lr){4-5}
		\cmidrule(lr){6-7}
		
		\# Labels per Class 
		& 2 & 4 
		& 4 & 10 
		& 2 & 4 \\
		\midrule
		
		FreeMatch (200k iters)
		& 78.46 & 83.58
		& 87.20 & 91.51
		& 93.48 & 94.20 \\

		  FreePiS (147k iters)
		& \textbf{80.61}
		& \textbf{84.27}
		& \textbf{90.44} 
		& \textbf{92.85}
		& \textbf{94.70} 
		& \textbf{95.84}  \\
        
		  \textcolor{gray!60}{FreePiS (200k iters)}
		& \textcolor{gray!60}{80.83} 
		& \textcolor{gray!60}{84.45}
		& \textcolor{gray!60}{90.65} 
		& \textcolor{gray!60}{93.13}
		& \textcolor{gray!60}{94.92}
		& \textcolor{gray!60}{95.89}  \\
		
		\bottomrule
	\end{tabular}
\end{table}

\section{Absolute Performance under Varying Data Scales}
\label{appendix:absolute_data_results}

This section reports the exact sample counts, data ratios, and absolute classification accuracy corresponding to the experiments shown in Figure~\ref{fig:combined}(a--b).

\subsection{Performance with Fixed Labeled Samples (\texorpdfstring{$N_l = 10$}{Nl = 10})}
In this setting, the number of labeled samples is fixed at 10 per class, while the amount of unlabeled data ($N_u$) is progressively increased. The unlabeled data ratio is defined as $N_u / (N_l + N_u)$. Table~\ref{tab:abs_fixed_labeled} summarizes the absolute performance under these conditions.

\begin{table}[ht]
\centering
\caption{Absolute classification accuracy (\%) with a fixed labeled count ($N_l = 10$). $^\ddagger$ denotes that parameter update sharing is enabled.}
\label{tab:abs_fixed_labeled}
\begin{tabular}{@{}lcccccc@{}}
\toprule
\multirow{2}{*}{Method}
& 9\%$^{\ddagger}$ & 50\%$^{\ddagger}$ & 71\% & 80\% & 85\% \\ 
& \textbf{($N_u=1$)} & \textbf{($N_u=10$)} & \textbf{($N_u=25$)} & \textbf{($N_u=40$)} & \textbf{($N_u=56$)} \\ \midrule
FreeMatch      & 56.45 & 59.88 & 63.01 & 65.66 & 65.87 \\
RegMixMatch    & \textbf{56.86} & 60.13 & 63.55 & 66.19 & 66.40 \\
\textbf{FreePiS} & 56.56 & \textbf{61.24} & \textbf{64.72} & \textbf{68.35} & \textbf{70.33} \\ \bottomrule
\end{tabular}
\end{table}

\subsection{Performance with Fixed Unlabeled Samples (\texorpdfstring{$N_u = 33$}{Nu = 33})}
In this setting, the number of unlabeled samples is fixed at 33 per class, while the labeled data volume ($N_l$) varies. The labeled data ratio is defined as $N_l / (N_l + N_u)$. Table~\ref{tab:abs_fixed_unlabeled} reports the corresponding results.

\begin{table}[ht]
\centering
\caption{Absolute classification accuracy (\%) with a fixed unlabeled count ($N_u = 33$). $^\ddagger$ denotes that parameter update sharing is enabled.}
\label{tab:abs_fixed_unlabeled}
\begin{tabular}{@{}lccccc@{}}
\toprule
\multirow{2}{*}{Method}
& 50\%$^{\ddagger}$ & 38\%$^{\ddagger}$ & 31\%$^{\ddagger}$ & 23\% & 13\% \\ 
& \textbf{($N_l=33$)} & \textbf{($N_l=20$)} & \textbf{($N_l=15$)} & \textbf{($N_l=10$)} & \textbf{($N_l=5$)} \\ \midrule
FreeMatch      & 83.34 & 74.77 & 70.39 & 62.87 & 49.65 \\
RegMixMatch    & 84.10 & 75.24 & 71.27 & 63.40 & 47.89 \\
\textbf{FreePiS} & \textbf{84.56} & \textbf{76.26} & \textbf{73.11} & \textbf{67.33} & \textbf{54.54} \\ \bottomrule
\end{tabular}
\end{table}

\section{Tabulated Results for Reliability and Accuracy}
\label{app:tab_fgd}

For completeness, we report the pseudo-label reliability and classification accuracy shown in Figure~\ref{fig:combined}(c--d) in tabular form. The corresponding numerical results are summarized in Table~\ref{tab:tab_fgd}.

We note that the numerical results reported in Table~\ref{tab:tab_fgd} (and Figure~\ref{fig:combined}(c--d)) are obtained from a single run with a fixed random seed (seed0) on CIFAR-100 with 2 labels per class. In contrast, the main results reported in Table~\ref{tab:pis+ssl} are averaged over three independent runs with different random seeds (seed0, seed1, seed2), following standard evaluation protocols.

As can be clearly observed, for FreeMatch, increasing the generalization intensity leads to a substantial degradation in the reliability of teacher-generated pseudo-labels. These unreliable pseudo-labels misguide the training process, ultimately resulting in a pronounced decline in classification accuracy.

\begin{table}[ht]
\centering
\caption{Tabulated results for classification accuracy and pseudo-label reliability under varying augmentation intensities $N$ and dropout rates.}
\label{tab:tab_fgd}
\resizebox{\textwidth}{!}{
\begin{tabular}{@{}llccccccccccc@{}}
\toprule
\multirow{2}{*}{Metric} & \multirow{2}{*}{Method}
& \multicolumn{6}{c}{Augmentation Intensity $N$ ($M=2N$)} 
& \multicolumn{5}{c}{Dropout Rate} \\ \cmidrule(lr){3-8} \cmidrule(lr){9-13}
& 
& 3 & 4 & 5 & 6 & 7 & 8
& 0.10 & 0.15 & 0.20 & 0.25 & 0.30 \\ \midrule

\multirow{2}{*}{Accuracy (\%)} 
& FreeMatch       
& 80.44 & 80.02 & 78.65 & 77.96 & 77.00 & 76.65
& 80.60 & 80.71 & 80.44 & 79.08 & 77.98 \\
& \textbf{FreePiS} 
& \textbf{80.91} & \textbf{81.66} & \textbf{82.18} & \textbf{82.03} & \textbf{81.11} & \textbf{80.80}
& \textbf{81.49} & \textbf{82.06} & \textbf{82.18} & \textbf{81.23} & \textbf{80.77} \\ \midrule

\multirow{2}{*}{Reliability (\%)} 
& FreeMatch       
& 79.24 & 78.09 & 76.19 & 74.45 & 73.17 & 72.68
& 79.95 & 80.01 & 79.24 & 77.09 & 75.35 \\
& \textbf{FreePiS} 
& \textbf{80.22} & \textbf{81.37} & \textbf{82.67} & \textbf{82.22} & \textbf{81.39} & \textbf{80.36}
& \textbf{81.36} & \textbf{82.53} & \textbf{82.67} & \textbf{81.66} & \textbf{81.00} \\ \bottomrule
\end{tabular}
}
\end{table}

\section{Synchronization Mechanism Ablation}
\label{appendix:sync}

The periodic parameter overwrite in PiS transfers the pioneer student's parameters
to the T--S model via a hard replacement: $\theta_s \leftarrow \theta_\pi$.
While conceptually simple, this design choice is deliberate.
In this appendix, we compare hard overwrite against one family of softer synchronization
mechanisms to justify this design and illuminate why a hard replacement is essential for
escaping labeled-dominated local minima.

\subsection{Synchronization Mechanisms}

We evaluate the following synchronization strategies, all applied at the same stage boundaries
as the original hard overwrite (i.e., every $\rho = 2.5\%$ of total training progress,
corresponding to 5{,}000 iterations in our standard 200{,}000-iteration training schedule).

\paragraph{Hard Overwrite (HO).}
The original mechanism. At the end of each stage, the T--S student parameters are
\emph{entirely replaced} by the pioneer student parameters:
\begin{equation}
    \theta_s \leftarrow \theta_\pi.
    \label{eq:hard_overwrite}
\end{equation}
This constitutes a one-shot, complete transfer of the pioneer student's accumulated knowledge,
with no retention of the T--S student's prior state.

\paragraph{Linear Interpolation (LI-$\alpha$).}
At the end of each stage, the T--S student parameters are updated as a convex combination
of the current T--S student and the pioneer student:
\begin{equation}
    \theta_s \leftarrow (1 - \alpha)\,\theta_s + \alpha\,\theta_\pi, \quad \alpha \in (0, 1].
    \label{eq:linear_interp}
\end{equation}
When $\alpha = 1$, this reduces to Hard Overwrite. Smaller $\alpha$ values retain more of the
T--S student's prior state, producing a smoother but less decisive parameter update.
We evaluate $\alpha \in \{0.3, 0.5, 0.7\}$.

\subsection{Results and Analysis}

\begin{table}[h]
\centering
\caption{
    Classification accuracy (\%) on CIFAR-100 under different
    synchronization mechanisms. All PiS variants use the same stage proportion $\rho = 2.5\%$ and warm-up $\phi = 8\%$.
}
\label{tab:sync_ablation}
\begin{tabular}{llc}
\toprule
Method & Sync Mechanism & Accuracy (\%) \\
\midrule
FreeMatch & --- & 78.46 \\
\midrule
FreePiS + LI-$\alpha$=0.3 & Linear Interp.\ ($\alpha$=0.3) & 78.80 \\
FreePiS + LI-$\alpha$=0.5 & Linear Interp.\ ($\alpha$=0.5) & 79.34 \\
FreePiS + LI-$\alpha$=0.7 & Linear Interp.\ ($\alpha$=0.7) & 80.19 \\
\midrule
\textbf{FreePiS + HO} & \textbf{Hard Overwrite ($\alpha$=1.0)} & \textbf{80.83} \\
\bottomrule
\end{tabular}
\end{table}

All experiments are conducted on CIFAR-100 with 2 labeled samples per class using FreePiS. As shown in Table~\ref{tab:sync_ablation}, hard overwrite consistently outperforms all
softer synchronization alternatives.
A key observation emerges from these results.

\textbf{Monotonic improvement with $\alpha$ in Linear Interpolation.}
Performance increases monotonically with $\alpha$, reaching its peak at $\alpha = 1.0$
(i.e., Hard Overwrite).
This monotonic relationship directly supports our interpretation of the overwrite as a
\emph{parameter-space teleportation}: the more completely the T--S student's parameters are
replaced by the pioneer student's, the more decisively the model escapes the
labeled-dominated local minimum in which it has been trapped.
A partial replacement ($\alpha < 1$) retains a residual component of the trapped parameters,
partially preserving the very sharp minimum that PiS is designed to escape.

These results suggest that the seeming simplicity of hard overwrite is in fact its strength. The abrupt and complete replacement enables the T–S model to escape labeled-dominated local minima, which smooth and partial blending mechanisms fundamentally fail to achieve.

\section{Assets and Licensing}
\label{sec:appendix_assets}

In compliance with the NeurIPS checklist, we provide a comprehensive summary of the datasets and codebases used in this work. All assets are used strictly for non-commercial academic research in accordance with their respective usage terms.

\begin{table}[h]
    \centering
    \caption{Summary of datasets and codebases with usage terms.}
    \label{tab:assets_summary}
    \small
    \resizebox{\textwidth}{!}{
    \begin{tabular}{@{}lllll@{}}
        \toprule
        \textbf{Asset} & \textbf{Modality} & \textbf{Type} & \textbf{License / Terms} & \textbf{Reference} \\ \midrule

        \textbf{Datasets} & & & & \\
        CIFAR-10/100 & CV & Dataset & Non-commercial research use & \cite{krizhevsky2009learning} \\
        STL-10 & CV & Dataset & Non-commercial research use & \cite{coates2011analysis} \\
        ImageNet-1K & CV & Dataset & ImageNet Terms of Access (non-commercial use) & \cite{deng2009imagenet} \\
        EuroSAT & CV & Dataset & MIT License & \cite{helber2019eurosat} \\
        FGVC-Aircraft & CV & Dataset & Non-commercial research use & \cite{fgvc} \\
        UrbanSound8K & Audio & Dataset & CC BY-NC-ND 4.0 & \cite{salamon2014dataset} \\
        ACL-IMDb & NLP & Dataset & Released for research use (non-commercial) & \cite{maas2011learning} \\ \midrule

        \textbf{Codebases} & & & & \\
        USB & Framework & Codebase & MIT License & \cite{wang2022usb} \\
        FixMatch & Algorithm & Codebase & Open-source implementation (license varies by repository) & \cite{sohn2020fixmatch} \\
        FreeMatch & Algorithm & Codebase & MIT License & \cite{wang2023freematch} \\
        RegMixMatch & Algorithm & Codebase & Open-source implementation (MIT-style license) & \cite{han2025regmixmatch} \\
        \bottomrule
    \end{tabular}
    }
\end{table}

\paragraph{Dataset Usage Details}
\begin{itemize}
    \item \textbf{Computer Vision (CV)}: CIFAR-10/100 and STL-10 are standard benchmarks widely used in academic research under non-commercial usage conditions. EuroSAT is distributed under the MIT License. FGVC-Aircraft is used under its original non-commercial research agreement. ImageNet-1K is used under the official ImageNet Terms of Access.
    
    \item \textbf{Audio}: UrbanSound8K is used under the Creative Commons Attribution-NonCommercial-NoDerivatives 4.0 International (CC BY-NC-ND 4.0) license.
    
    \item \textbf{NLP}: The ACL-IMDb dataset is used under its original research-use-only release conditions.
\end{itemize}

\paragraph{Software and Codebases}
\begin{itemize}
    \item \textbf{USB (Unified SSL Benchmark)}: Built upon the USB framework, which is released under the MIT License.
    
    \item \textbf{Baseline Implementations}: We use official or widely adopted open-source implementations for all baseline methods. These implementations follow their respective repository licenses, typically MIT or similar permissive licenses.
\end{itemize}


\newpage
\section*{NeurIPS Paper Checklist}

\begin{enumerate}

\item {\bf Claims}
    \item[] Question: Do the main claims made in the abstract and introduction accurately reflect the paper's contributions and scope?
    \item[] Answer: \answerYes{} 
    \item[] Justification: We clearly state the claims regarding the Fitting-Generalization Dilemma (FGD) and Label-Driven Trap (LDT) in the abstract and introduction, which strictly match our theoretical analysis and experimental results using the Pioneer Student (PiS) branch.
    \item[] Guidelines:
    \begin{itemize}
        \item The answer \answerNA{} means that the abstract and introduction do not include the claims made in the paper.
        \item The abstract and/or introduction should clearly state the claims made, including the contributions made in the paper and important assumptions and limitations. A \answerNo{} or \answerNA{} answer to this question will not be perceived well by the reviewers. 
        \item The claims made should match theoretical and experimental results, and reflect how much the results can be expected to generalize to other settings. 
        \item It is fine to include aspirational goals as motivation as long as it is clear that these goals are not attained by the paper. 
    \end{itemize}

\item {\bf Limitations}
    \item[] Question: Does the paper discuss the limitations of the work performed by the authors?
    \item[] Answer: \answerYes{} 
    \item[] Justification: We discuss the performance trade-offs in high-label scenarios (Section~\ref{sec:when_excel}) that necessitate the parameter update sharing mechanism, as well as the increased single-step computational overhead (Section~\ref{sec:conclusion}) introduced by the dual-branch architecture.
    \item[] Guidelines:
    \begin{itemize}
        \item The answer \answerNA{} means that the paper has no limitation while the answer \answerNo{} means that the paper has limitations, but those are not discussed in the paper. 
        \item The authors are encouraged to create a separate ``Limitations'' section in their paper.
        \item The paper should point out any strong assumptions and how robust the results are to violations of these assumptions (e.g., independence assumptions, noiseless settings, model well-specification, asymptotic approximations only holding locally). The authors should reflect on how these assumptions might be violated in practice and what the implications would be.
        \item The authors should reflect on the scope of the claims made, e.g., if the approach was only tested on a few datasets or with a few runs. In general, empirical results often depend on implicit assumptions, which should be articulated.
        \item The authors should reflect on the factors that influence the performance of the approach. For example, a facial recognition algorithm may perform poorly when image resolution is low or images are taken in low lighting. Or a speech-to-text system might not be used reliably to provide closed captions for online lectures because it fails to handle technical jargon.
        \item The authors should discuss the computational efficiency of the proposed algorithms and how they scale with dataset size.
        \item If applicable, the authors should discuss possible limitations of their approach to address problems of privacy and fairness.
        \item While the authors might fear that complete honesty about limitations might be used by reviewers as grounds for rejection, a worse outcome might be that reviewers discover limitations that aren't acknowledged in the paper. The authors should use their best judgment and recognize that individual actions in favor of transparency play an important role in developing norms that preserve the integrity of the community. Reviewers will be specifically instructed to not penalize honesty concerning limitations.
    \end{itemize}

\item {\bf Theory assumptions and proofs}
    \item[] Question: For each theoretical result, does the paper provide the full set of assumptions and a complete (and correct) proof?
    \item[] Answer: \answerNA{} 
    \item[] Justification: This paper does not present formal theoretical results such as theorems or proofs. We introduce the Update Consistency (UC) metric as a formal definition and provide empirical analysis, including loss landscape visualization, to support our claims regarding the Label-Driven Trap (LDT).
    \item[] Guidelines:
    \begin{itemize}
        \item The answer \answerNA{} means that the paper does not include theoretical results. 
        \item All the theorems, formulas, and proofs in the paper should be numbered and cross-referenced.
        \item All assumptions should be clearly stated or referenced in the statement of any theorems.
        \item The proofs can either appear in the main paper or the supplemental material, but if they appear in the supplemental material, the authors are encouraged to provide a short proof sketch to provide intuition. 
        \item Inversely, any informal proof provided in the core of the paper should be complemented by formal proofs provided in appendix or supplemental material.
        \item Theorems and Lemmas that the proof relies upon should be properly referenced. 
    \end{itemize}

    \item {\bf Experimental result reproducibility}
    \item[] Question: Does the paper fully disclose all the information needed to reproduce the main experimental results of the paper to the extent that it affects the main claims and/or conclusions of the paper (regardless of whether the code and data are provided or not)?
    \item[] Answer: \answerYes{} 
    \item[] Justification: We provide full architectural details of the plug-and-play PiS module (Section~\ref{sec:Method}). We also include key experimental configurations in Appendix~\ref{appendix:training_configs}.
    \item[] Guidelines:
    \begin{itemize}
        \item The answer \answerNA{} means that the paper does not include experiments.
        \item If the paper includes experiments, a \answerNo{} answer to this question will not be perceived well by the reviewers: Making the paper reproducible is important, regardless of whether the code and data are provided or not.
        \item If the contribution is a dataset and\slash or model, the authors should describe the steps taken to make their results reproducible or verifiable. 
        \item Depending on the contribution, reproducibility can be accomplished in various ways. For example, if the contribution is a novel architecture, describing the architecture fully might suffice, or if the contribution is a specific model and empirical evaluation, it may be necessary to either make it possible for others to replicate the model with the same dataset, or provide access to the model. In general. releasing code and data is often one good way to accomplish this, but reproducibility can also be provided via detailed instructions for how to replicate the results, access to a hosted model (e.g., in the case of a large language model), releasing of a model checkpoint, or other means that are appropriate to the research performed.
        \item While NeurIPS does not require releasing code, the conference does require all submissions to provide some reasonable avenue for reproducibility, which may depend on the nature of the contribution. For example
        \begin{enumerate}
            \item If the contribution is primarily a new algorithm, the paper should make it clear how to reproduce that algorithm.
            \item If the contribution is primarily a new model architecture, the paper should describe the architecture clearly and fully.
            \item If the contribution is a new model (e.g., a large language model), then there should either be a way to access this model for reproducing the results or a way to reproduce the model (e.g., with an open-source dataset or instructions for how to construct the dataset).
            \item We recognize that reproducibility may be tricky in some cases, in which case authors are welcome to describe the particular way they provide for reproducibility. In the case of closed-source models, it may be that access to the model is limited in some way (e.g., to registered users), but it should be possible for other researchers to have some path to reproducing or verifying the results.
        \end{enumerate}
    \end{itemize}

\item {\bf Open access to data and code}
    \item[] Question: Does the paper provide open access to the data and code, with sufficient instructions to faithfully reproduce the main experimental results, as described in supplemental material?
    \item[] Answer: \answerYes{} 
    \item[] Justification: We include the anonymized source code, environment dependencies, and executable scripts for reproducing the experimental results in the supplemental material.
    \item[] Guidelines:
    \begin{itemize}
        \item The answer \answerNA{} means that paper does not include experiments requiring code.
        \item Please see the NeurIPS code and data submission guidelines (\url{https://neurips.cc/public/guides/CodeSubmissionPolicy}) for more details.
        \item While we encourage the release of code and data, we understand that this might not be possible, so \answerNo{} is an acceptable answer. Papers cannot be rejected simply for not including code, unless this is central to the contribution (e.g., for a new open-source benchmark).
        \item The instructions should contain the exact command and environment needed to run to reproduce the results. See the NeurIPS code and data submission guidelines (\url{https://neurips.cc/public/guides/CodeSubmissionPolicy}) for more details.
        \item The authors should provide instructions on data access and preparation, including how to access the raw data, preprocessed data, intermediate data, and generated data, etc.
        \item The authors should provide scripts to reproduce all experimental results for the new proposed method and baselines. If only a subset of experiments are reproducible, they should state which ones are omitted from the script and why.
        \item At submission time, to preserve anonymity, the authors should release anonymized versions (if applicable).
        \item Providing as much information as possible in supplemental material (appended to the paper) is recommended, but including URLs to data and code is permitted.
    \end{itemize}

\item {\bf Experimental setting/details}
    \item[] Question: Does the paper specify all the training and test details (e.g., data splits, hyperparameters, how they were chosen, type of optimizer) necessary to understand the results?
    \item[] Answer: \answerYes{} 
    \item[] Justification: We detail the specific hyperparameter configurations introduced by PiS (e.g., parameter overwrite interval $\rho=2.5\%$, warmup ratio $\phi=8.0\%$, update sharing strength $\beta=0.6$, see Section~\ref{sec:Experiments}), as well as standard SSL baseline settings (Appendix~\ref{appendix:training_configs}).
    \item[] Guidelines:
    \begin{itemize}
        \item The answer \answerNA{} means that the paper does not include experiments.
        \item The experimental setting should be presented in the core of the paper to a level of detail that is necessary to appreciate the results and make sense of them.
        \item The full details can be provided either with the code, in appendix, or as supplemental material.
    \end{itemize}

\item {\bf Experiment statistical significance}
    \item[] Question: Does the paper report error bars suitably and correctly defined or other appropriate information about the statistical significance of the experiments?
    \item[] Answer: \answerYes{} 
    \item[] Justification: We report the mean and standard deviation over three independent runs with different random seeds for all main results (see Tables~\ref{tab:pis+ssl} in Section~\ref{sec:Experiments}). The variability arises from random initialization and data sampling, and the reported error bars correspond to standard deviation.
    \item[] Guidelines:
    \begin{itemize}
        \item The answer \answerNA{} means that the paper does not include experiments.
        \item The authors should answer \answerYes{} if the results are accompanied by error bars, confidence intervals, or statistical significance tests, at least for the experiments that support the main claims of the paper.
        \item The factors of variability that the error bars are capturing should be clearly stated (for example, train/test split, initialization, random drawing of some parameter, or overall run with given experimental conditions).
        \item The method for calculating the error bars should be explained (closed form formula, call to a library function, bootstrap, etc.)
        \item The assumptions made should be given (e.g., Normally distributed errors).
        \item It should be clear whether the error bar is the standard deviation or the standard error of the mean.
        \item It is OK to report 1-sigma error bars, but one should state it. The authors should preferably report a 2-sigma error bar than state that they have a 96\% CI, if the hypothesis of Normality of errors is not verified.
        \item For asymmetric distributions, the authors should be careful not to show in tables or figures symmetric error bars that would yield results that are out of range (e.g., negative error rates).
        \item If error bars are reported in tables or plots, the authors should explain in the text how they were calculated and reference the corresponding figures or tables in the text.
    \end{itemize}

\item {\bf Experiments compute resources}
    \item[] Question: For each experiment, does the paper provide sufficient information on the computer resources (type of compute workers, memory, time of execution) needed to reproduce the experiments?
    \item[] Answer: \answerYes{} 
    \item[] Justification: We provide the compute resources used in our experiments, including GPU type (RTX 3090 with 24GB memory) and training time per run (12–20 hours), as described in Appendix~\ref{appendix:training_configs}.
    \item[] Guidelines:
    \begin{itemize}
        \item The answer \answerNA{} means that the paper does not include experiments.
        \item The paper should indicate the type of compute workers CPU or GPU, internal cluster, or cloud provider, including relevant memory and storage.
        \item The paper should provide the amount of compute required for each of the individual experimental runs as well as estimate the total compute. 
        \item The paper should disclose whether the full research project required more compute than the experiments reported in the paper (e.g., preliminary or failed experiments that didn't make it into the paper). 
    \end{itemize}
    
\item {\bf Code of ethics}
    \item[] Question: Does the research conducted in the paper conform, in every respect, with the NeurIPS Code of Ethics \url{https://neurips.cc/public/EthicsGuidelines}?
    \item[] Answer: \answerYes{} 
    \item[] Justification: This fundamental algorithmic research strictly adheres to the NeurIPS Code of Ethics and does not raise ethical concerns.
    \item[] Guidelines:
    \begin{itemize}
        \item The answer \answerNA{} means that the authors have not reviewed the NeurIPS Code of Ethics.
        \item If the authors answer \answerNo, they should explain the special circumstances that require a deviation from the Code of Ethics.
        \item The authors should make sure to preserve anonymity (e.g., if there is a special consideration due to laws or regulations in their jurisdiction).
    \end{itemize}

\item {\bf Broader impacts}
    \item[] Question: Does the paper discuss both potential positive societal impacts and negative societal impacts of the work performed?
    \item[] Answer: \answerNA{} 
    \item[] Justification: This work focuses on foundational optimization theory and architectural designs for semi-supervised learning; there are no foreseeable, direct negative societal impacts tied to specific applications.
    \item[] Guidelines:
    \begin{itemize}
        \item The answer \answerNA{} means that there is no societal impact of the work performed.
        \item If the authors answer \answerNA{} or \answerNo, they should explain why their work has no societal impact or why the paper does not address societal impact.
        \item Examples of negative societal impacts include potential malicious or unintended uses (e.g., disinformation, generating fake profiles, surveillance), fairness considerations (e.g., deployment of technologies that could make decisions that unfairly impact specific groups), privacy considerations, and security considerations.
        \item The conference expects that many papers will be foundational research and not tied to particular applications, let alone deployments. However, if there is a direct path to any negative applications, the authors should point it out. For example, it is legitimate to point out that an improvement in the quality of generative models could be used to generate Deepfakes for disinformation. On the other hand, it is not needed to point out that a generic algorithm for optimizing neural networks could enable people to train models that generate Deepfakes faster.
        \item The authors should consider possible harms that could arise when the technology is being used as intended and functioning correctly, harms that could arise when the technology is being used as intended but gives incorrect results, and harms following from (intentional or unintentional) misuse of the technology.
        \item If there are negative societal impacts, the authors could also discuss possible mitigation strategies (e.g., gated release of models, providing defenses in addition to attacks, mechanisms for monitoring misuse, mechanisms to monitor how a system learns from feedback over time, improving the efficiency and accessibility of ML).
    \end{itemize}
    
\item {\bf Safeguards}
    \item[] Question: Does the paper describe safeguards that have been put in place for responsible release of data or models that have a high risk for misuse (e.g., pre-trained language models, image generators, or scraped datasets)?
    \item[] Answer: \answerNA{} 
    \item[] Justification: Our contribution is a general optimization framework; we do not release any high-risk assets, large pre-trained generative models, or newly scraped datasets.
    \item[] Guidelines:
    \begin{itemize}
        \item The answer \answerNA{} means that the paper poses no such risks.
        \item Released models that have a high risk for misuse or dual-use should be released with necessary safeguards to allow for controlled use of the model, for example by requiring that users adhere to usage guidelines or restrictions to access the model or implementing safety filters. 
        \item Datasets that have been scraped from the Internet could pose safety risks. The authors should describe how they avoided releasing unsafe images.
        \item We recognize that providing effective safeguards is challenging, and many papers do not require this, but we encourage authors to take this into account and make a best faith effort.
    \end{itemize}

\item {\bf Licenses for existing assets}
    \item[] Question: Are the creators or original owners of assets (e.g., code, data, models), used in the paper, properly credited and are the license and terms of use explicitly mentioned and properly respected?
    \item[] Answer: \answerYes{} 
    \item[] Justification: We properly cite all datasets and codebases used in this work and explicitly specify their licenses (see Appendix~\ref{sec:appendix_assets} for a complete list of asset licenses).
    \item[] Guidelines:
    \begin{itemize}
        \item The answer \answerNA{} means that the paper does not use existing assets.
        \item The authors should cite the original paper that produced the code package or dataset.
        \item The authors should state which version of the asset is used and, if possible, include a URL.
        \item The name of the license (e.g., CC-BY 4.0) should be included for each asset.
        \item For scraped data from a particular source (e.g., website), the copyright and terms of service of that source should be provided.
        \item If assets are released, the license, copyright information, and terms of use in the package should be provided. For popular datasets, \url{paperswithcode.com/datasets} has curated licenses for some datasets. Their licensing guide can help determine the license of a dataset.
        \item For existing datasets that are re-packaged, both the original license and the license of the derived asset (if it has changed) should be provided.
        \item If this information is not available online, the authors are encouraged to reach out to the asset's creators.
    \end{itemize}

\item {\bf New assets}
    \item[] Question: Are new assets introduced in the paper well documented and is the documentation provided alongside the assets?
    \item[] Answer: \answerNA{} 
    \item[] Justification: The paper proposes a novel optimization algorithm and architectural module; it does not introduce new standalone datasets or similar assets requiring independent documentation.
    \item[] Guidelines:
    \begin{itemize}
        \item The answer \answerNA{} means that the paper does not release new assets.
        \item Researchers should communicate the details of the dataset\slash code\slash model as part of their submissions via structured templates. This includes details about training, license, limitations, etc. 
        \item The paper should discuss whether and how consent was obtained from people whose asset is used.
        \item At submission time, remember to anonymize your assets (if applicable). You can either create an anonymized URL or include an anonymized zip file.
    \end{itemize}

\item {\bf Crowdsourcing and research with human subjects}
    \item[] Question: For crowdsourcing experiments and research with human subjects, does the paper include the full text of instructions given to participants and screenshots, if applicable, as well as details about compensation (if any)? 
    \item[] Answer: \answerNA{} 
    \item[] Justification: The research relies solely on computational evaluations using standard academic benchmarks and does not involve any human subjects or crowdsourcing.
    \item[] Guidelines:
    \begin{itemize}
        \item The answer \answerNA{} means that the paper does not involve crowdsourcing nor research with human subjects.
        \item Including this information in the supplemental material is fine, but if the main contribution of the paper involves human subjects, then as much detail as possible should be included in the main paper. 
        \item According to the NeurIPS Code of Ethics, workers involved in data collection, curation, or other labor should be paid at least the minimum wage in the country of the data collector. 
    \end{itemize}

\item {\bf Institutional review board (IRB) approvals or equivalent for research with human subjects}
    \item[] Question: Does the paper describe potential risks incurred by study participants, whether such risks were disclosed to the subjects, and whether Institutional Review Board (IRB) approvals (or an equivalent approval/review based on the requirements of your country or institution) were obtained?
    \item[] Answer: \answerNA{} 
    \item[] Justification: The research does not involve human subjects, making IRB approval inapplicable.
    \item[] Guidelines:
    \begin{itemize}
        \item The answer \answerNA{} means that the paper does not involve crowdsourcing nor research with human subjects.
        \item Depending on the country in which research is conducted, IRB approval (or equivalent) may be required for any human subjects research. If you obtained IRB approval, you should clearly state this in the paper. 
        \item We recognize that the procedures for this may vary significantly between institutions and locations, and we expect authors to adhere to the NeurIPS Code of Ethics and the guidelines for their institution. 
        \item For initial submissions, do not include any information that would break anonymity (if applicable), such as the institution conducting the review.
    \end{itemize}

\item {\bf Declaration of LLM usage}
    \item[] Question: Does the paper describe the usage of LLMs if it is an important, original, or non-standard component of the core methods in this research? Note that if the LLM is used only for writing, editing, or formatting purposes and does \emph{not} impact the core methodology, scientific rigor, or originality of the research, declaration is not required.
    \item[] Answer: \answerNA{} 
    \item[] Justification: Large Language Models (LLMs) are not utilized as a component of the core Pioneer Student methodology or the semi-supervised learning algorithms proposed.
    \item[] Guidelines:
    \begin{itemize}
        \item The answer \answerNA{} means that the core method development in this research does not involve LLMs as any important, original, or non-standard components.
        \item Please refer to our LLM policy in the NeurIPS handbook for what should or should not be described.
    \end{itemize}

\end{enumerate}

\end{document}